\documentclass[journal]{IEEEtran}
\usepackage[utf8]{inputenc} 
\usepackage[T1]{fontenc}    
\usepackage{hyperref}       
\usepackage{latexsym,bm}
\usepackage[tbtags]{amsmath}
\usepackage{graphicx}
\usepackage{subfigure}
\usepackage{flushend}
\usepackage{amssymb}
\usepackage{stfloats}
\usepackage{cite}
\usepackage{stfloats}
\usepackage{comment}
\usepackage{algorithm}
\usepackage{algpseudocode}
\usepackage{array}
\usepackage{amsmath}
\usepackage{bm}
\usepackage{CJK}
\usepackage{setspace}
\usepackage{algorithm}
\usepackage{color}
\usepackage{verbatim}
\usepackage{multirow}
\usepackage{url}
\usepackage{amssymb}
\newcommand{\Cov}{\operatorname{Cov}}
\begin{document}

\title{SPTTE: A Spatiotemporal Probabilistic Framework for Travel Time Estimation}

\author{
Chen~Xu,
Qiang~Wang, \emph{Member, IEEE} and 
Lijun~Sun, \emph{Senior Member, IEEE}

\thanks{
We acknowledge the support of the Natural Sciences and Engineering Research Council of Canada (NSERC) through the Discovery Grant.

Chen~Xu and Qiang~Wang are with the National Engineering Research Center of Mobile Network Technologies, Beijing University of Posts and Telecommunications, Beijing, China (e-mail: xuchen; wangq@bupt.edu.cn).

Lijun Sun is with the Department of Civil Engineering, McGill University, Montreal, Quebec, Canada (e-mail: lijun.sun@mcgil.ca).

\textit{(Corresponding author: Lijun Sun.)}
}
}

\maketitle
\begin{abstract}
Accurate travel time estimation is essential for navigation and itinerary planning. While existing research employs probabilistic modeling to assess travel time uncertainty and account for correlations between multiple trips, modeling the temporal variability of multi-trip travel time distributions remains a significant challenge. Capturing the evolution of joint distributions requires large, well-organized datasets; however, real-world trip data are often temporally sparse and spatially unevenly distributed. To address this issue, we propose SPTTE, a spatiotemporal probabilistic framework that models the evolving joint distribution of multi-trip travel times by formulating the estimation task as a spatiotemporal stochastic process regression problem with fragmented observations. SPTTE incorporates an RNN-based temporal Gaussian process parameterization to regularize sparse observations and capture temporal dependencies. Additionally, it employs a prior-based heterogeneity smoothing strategy to correct unreliable learning caused by unevenly distributed trips, effectively modeling temporal variability under sparse and uneven data distributions. Evaluations on real-world datasets demonstrate that SPTTE outperforms state-of-the-art deterministic and probabilistic methods by over 10.13\%. Ablation studies and visualizations further confirm the effectiveness of the model components.
\end{abstract}


\IEEEpeerreviewmaketitle

\section{Introduction}

Travel time estimation (TTE) refers to the task of predicting the time required for a vehicle to travel along a queried path connecting two locations and is crucial for navigation and itinerary planning.  It can be conceptualized as a regression task within a spatiotemporal domain \cite{yan2024efficiency}, characterized by sparse, irregular, and fragmented observations \cite{chen2022bayesian}. Compared to traditional spatiotemporal data prediction (e.g., time series forecasting), TTE is more complex due to the temporal continuity and the intricate spatial dependencies inherent in travel sequences. Consequently, TTE has attracted considerable attention in recent years, providing essential data support for technologies such as autonomous vehicle decision-making and navigation.

Research on TTE primarily develops along two directions based on the availability of route information \cite{yan2024efficiency}. In the absence of route-specific information, TTE relies solely on origin-destination (OD) data as the primary reference \cite{wang2014travel,wang2019simple,li2018multitask,yuan2020effective,lin2023origindestination,jindal2017unified,wang2023multitask}. This OD-based methodology leverages historical travel times from trips with comparable OD spatial configurations to infer the travel time of the target trip \cite{li2018multitask,lin2023origindestination,wang2023multitask}. To enhance the model's representational capacity, additional factors such as departure time, travel distance, and road attributes \cite{yuan2020effective,jindal2017unified,wang2023multitask} are progressively incorporated into the feature space. While computationally efficient and requiring minimal data, this approach's reliance on OD information alone constrains its predictive accuracy. The absence of detailed route-specific data results in inherently imprecise estimates, leading to lower temporal and spatial resolution in travel time predictions.

The advancement and widespread adoption of high-precision GPS technology have made route information both reliable and easily accessible, fueling the rapid development of the second approach: route-based travel time prediction. This method utilizes detailed trip data to model the specific routes taken during travel. Unlike OD-based approaches, which rely on general spatial locations, route-based methods focus on learning representations of individual road segments \cite{han2021multisemantic,liao2024multifaceted}, GPS coordinates \cite{wang2018when}, or trajectory attributes associated with the route itself. This allows for the capture of more granular spatiotemporal dependencies, accounting for both temporal dynamics \cite{qiu2019neitte,wang2018learning} (e.g., traffic patterns over time) and spatial interactions \cite{hong2020heteta,fang2020constgat} (e.g., congestion on specific roads or intersections) that influence travel time along the route. By leveraging techniques such as recurrent neural networks (RNNs) \cite{sun2020codriver,ye2022cateta} for temporal dependencies and graph neural networks (GNNs) \cite{wang2021graphtte,derrow2021eta} for modeling spatial relationships, route-based methods provide more accurate and context-aware predictions, especially in complex, dynamic traffic environments. Additionally, techniques like attention mechanisms \cite{zou2023when,chen2022interpreting} and meta-learning \cite{wang2022finegrained,fang2021ssml} have further enriched these models' capabilities.

The deterministic estimation models mentioned above typically provide only the mean travel time, yet the confidence level of these estimates is often more important. Recent work \cite{li2019learning,james2021citywide,zhou2023travel} has increasingly focused on probabilistic estimation of travel time. A fundamental assumption in the majority of these studies is that travel times for different trips are independent. While this simplification facilitates modeling each trip instance individually and allows the model to be efficiently learned, it ignores the potential correlation for trips with overlapping links. Considering the correlations between multiple trips, our prior work \cite{xu2024link} proposed \textit{ProbETA}, a probabilistic deep learning model that can capture both inter-trip correlations and intra-trip correlations. \textit{ProbETA} leverages Gaussian parameterization to learn link representations, accounting for both time-specific and trip-specific random effects. However, \textit{ProbETA} is a static model in which the temporal patterns for a specific time window (e.g., morning peak hour) are only captured by combining data of different days. This approach requires extensive training for each predefined time window and also overlooks the time-varying nature of traffic states. 

Two substantial challenges persist in effectively modeling the temporal variability of travel time distributions: (1) The temporal sparsity of trips makes it impossible to observe the actual travel time of certain trips at all time slots, posing a fundamental challenge in capturing temporal dependencies; (2) The spatial unevenness of trips across the road network results in incomplete or coarse coverage of certain links, leading to effective learning in high-frequency trip areas while low-frequency areas remain challenging to optimize reliably.

\begin{figure}
	\centering
	\includegraphics[width=3in]{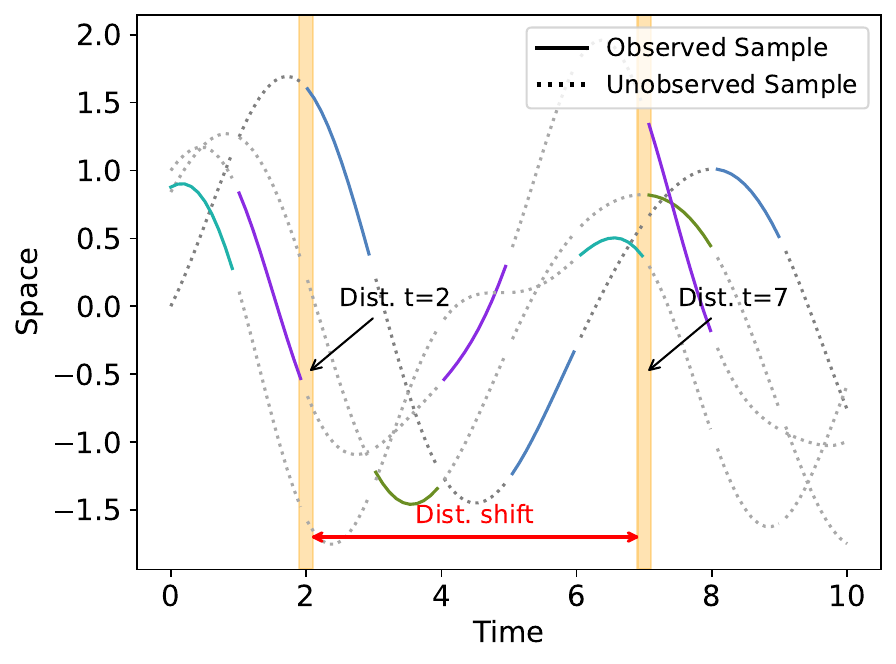}
	\caption{Fragmented realization of the trip spatiotemporal stochastic process. The space is simplified to one dimension for illustration. Each curve represents a realization of the trip stochastic process, with the solid line indicating the observed fragment and the dashed line representing the unobserved fragment.}
	\label{fig:rea}
\end{figure}

To address these challenges, we formulate the joint distribution estimation of multiple trip travel times as a fragmented realization-based \cite{delaigle2021estimating,lin2022mean} spatiotemporal stochastic process regression problem, as shown in Figure~\ref{fig:rea}, and propose a Spatiotemporal joint Probability model for Travel Time Estimation (\textit{SPTTE}). Specifically, we propose an RNN-based temporal Gaussian process parameterization to standardize the sparsely fragmented trip realizations using learnable link representations and capture the temporal dependencies of travel time distributions based on the coverage frequency of trips. Additionally, a heterogeneity smoothing strategy based on similarity derived from prior knowledge and frequency-based heterogeneous weights is developed to address unreliable learning due to uneven trip distribution. The main contributions are as follows:
\begin{itemize} 
\item We propose a spatiotemporal joint probability distribution estimation model for multi-trip travel times. A temporal Gaussian process parameterization method is designed to standardize trip representations and capture temporal dependencies based on the coverage frequency of historical trips.

\item A prior-based heterogeneity smoothing strategy is developed to address unreliable learning induced by uneven sample distributions. This strategy corrects unreliable link representations by applying asymmetric smoothing to neighborhoods, leveraging carefully designed heterogeneity weights and prior similarity.

\item We validate our model on the Harbin and Chengdu datasets, where it outperforms state-of-the-art baselines by more than 10.13\% relatively, and interpretably models temporal variations in travel time means and correlations. Ablation and visualization analyses further confirm the effectiveness of the learned evolution. 

\end{itemize}

The structure of this paper is as follows: Section~\ref{sec:related} reviews relevant literature, Section~\ref{sec:formulation} outlines foundational definitions and formulations, Section~\ref{sec:method} details the main components of the \textit{SPTTE} framework, Section~\ref{sec:results} provides an empirical evaluation using two real-world datasets, and finally, Section~\ref{sec:conclusion} concludes the paper with a summary of findings.

\section{Related Work} \label{sec:related}

\begin{table*}
	\centering
	\caption{Comparison of Travel Time Estimation Models.}
	\label{tab:review}
	\scriptsize
	\renewcommand {\arraystretch}{1.3}
	\resizebox{\textwidth}{31mm}{
		\begin{tabular}{c c c c c c c  } 
			\hline
			\hline
			{Model Name}& {Model Type}&Data Requirement&Probabilistic & Multi-trip Corr & Link Representation&Trip Representation \\
			\hline\hline
			{MURAT (2018)\cite{li2018multitask}  } &{OD-based } & {OD Pair + Road Attributes } & { }&{}& {Graph Laplacian Regularization}& {OD Link Representation Concat}\\
			\hline
			{TEMP (2019)\cite{wang2019simple}  } &{OD-based } & {OD Pair } & { }&{}&--- &{OD Tuple}\\
			\hline
			{DeepOD (2020)\cite{yuan2020effective}  } &{OD-based } & {OD Pair + Road Attributes } & { }&{}& {Link and Attribute Embedding}& {OD Link Representation Concat}\\
			\hline
			{DOT (2023)\cite{lin2023origindestination}  } &{OD-based } & {OD Pair } & { }&{}& {---}& {Masked Vision Transformer}\\
			\hline
			{MWSL-TTE (2023)\cite{wang2023multitask}  } &{OD-based } & {OD Pair + Road Attributes } & { }&{}& {Relational GCN}& {Direct Addition}\\
			\hline
			{DeepTTE (2018)\cite{wang2018when}  } &{Route-based } & {GPS Sequence + Trip Attributes } & { }&{}& {Geo-Conv}& {LSTM}\\
			\hline
			{WDR (2018)\cite{wang2018learning} } &{Route-based } & {GPS Sequence + Trip Attributes } & { }&{}& {Wide \& Deep Model}& {LSTM}\\
			\hline
			{HetETA (2020)\cite{hong2020heteta}  } &{Route-based } & {GPS Sequence + Road Attributes } & { }&{}& {Het-ChebNet}& {Gated CNN}\\
			\hline
			{STTE (2021)\cite{han2021multisemantic}  } &{Route-based } & {GPS Sequence + Road Attributes } & { }&{}& {Multi-Semantic Embedding}& {LSTM}\\
			\hline
			{CatETA (2022)\cite{ye2022cateta}  } &{Route-based } & {GPS Sequence + Road Attributes + Trip Attributes } & { }&{}& {Link and Attribute Embedding}& {BiGRU}\\
			\hline  
			{HierETA (2022)\cite{chen2022interpreting}  } &{Route-based } & {GPS Sequence + Road Attributes } & { }&{}& {Link-Intersection Embedding}& {BiLSTM}\\
			\hline 
			{MulT-TTE (2024)\cite{liao2024multifaceted} } &{Route-based } & {GPS Sequence + Road Attributes } & { }&{}& {Multi-Faceted Embedding}& {Transformer}\\
			\hline
			{DeepGTT (2019) \cite{li2019learning} } &{Route-based } & {GPS Sequence + Road Attributes } & {\checkmark }&{}& {Link and Speed Embedding}& {Link Representation Weighted Addition}\\
			\hline
			{RTAG (2023) \cite{zhou2023travel} } &{Route-based } & {GPS Sequence + Road Attributes + Trip Attributes } & {\checkmark }&{}& {Link and Attribute Embedding}& {Self-attention with Trip Subgraph}\\
			\hline
			{GMDNet (2023) \cite{mao2023gmdnet} } &{Route-based } & {GPS Sequence + Road Attributes } & {\checkmark }&{}& {Link and Attribute Embedding}& {Multi-head Self-attention with Position}\\
			\hline
			{ProbETA (2024) \cite{xu2024link} } &{Route-based} &{GPS Sequence } & {\checkmark }& {\checkmark }& {Link Embedding }& {Path-based Gaussian Aggregation}\\
			\hline
			\hline      
			{\textbf{\textit{SPTTE} (ours) } } &{\textbf{Route-based}} &{\textbf{GPS Sequence + Road Attributes} } & {\textbf{\checkmark} }& {\textbf{\checkmark} }& {\textbf{GRU + Heterogeneous GCN} }& {\textbf{Path-based Gaussian Aggregation}}\\
			\hline
	\end{tabular}}
\end{table*}

This section provides an overview of existing research on travel time estimation, covering both deterministic and probabilistic approaches. A comparative analysis of various methods is summarized in Table~\ref{tab:review}.

\subsection{Deterministic Methods for TTE}

Travel time estimation (TTE) methods can be broadly classified into two categories based on input data: Origin-Destination (OD)-based methods and route-based methods. OD-based methods rely primarily on OD pairs as input \cite{wang2019simple,li2018multitask,yuan2020effective,lin2023origindestination,jindal2017unified,wang2023multitask}. Wang et al. \cite{wang2019simple} proposed a neighbor-based approach that identifies trips with similar origins and destinations while considering dynamic traffic conditions. By aggregating the travel times of these neighboring trips, the model estimates the travel time for the target trip. Li et al. \cite{li2018multitask} introduced MURAT, a multi-task learning model that enhances the feature space by incorporating trip properties and spatiotemporal prior knowledge, generating comprehensive trip representations. Yuan et al. \cite{yuan2020effective} incorporated trajectory data during training to improve the matching of neighboring trips, using road segment and time slot embeddings to capture spatiotemporal characteristics. This approach, similar to distillation learning, can effectively enhance the feature representation of OD pairs. Similarly, Lin et al. \cite{lin2023origindestination} integrated trajectory data by dividing the city into grid cells and using a Masked Vision Transformer to model pixel-level correlations, introducing a diffusion model to encode both the OD pair and the trip's trajectory. More recently, Wang et al. \cite{wang2023multitask} explored probabilistic modeling within the OD framework by inferring transition probabilities between road segments and identifying the most probable route, estimating trip travel time by summing the travel times of all road segments along this route. While OD-based approaches are versatile and require minimal data, their reliance on OD information alone can limit predictive accuracy due to the potential ambiguity of OD inputs and the absence of detailed route-specific data, leading to lower temporal and spatial resolution in travel time predictions.

Route-based methods utilize detailed trip data to model specific routes, treating a trip as a sequence of links or GPS points. Advanced deep neural networks are applied to capture sequence features and provide travel time estimations. Wang et al. \cite{wang2018when} proposed an approach that embeds GPS points and their geographical context by combining geo-convolution and recurrent units to capture spatial and temporal dependencies. Liao et al. \cite{liao2024multifaceted} introduced a multi-faceted route representation learning framework, dividing each route into sequences of GPS coordinates, road segment attributes, and road segment IDs, processed using a transformer encoder to generate travel time estimates. However, GPS data's high dimensionality and unstable accuracy can cause issues with model convergence.

Some studies focus on mapping GPS points to the road network before processing trajectories. Wang et al. \cite{wang2018learning} framed the TTE problem as a regression task, proposing a Wide-Deep-Recurrent (WDR) model that integrates wide linear models, deep neural networks, and recurrent neural networks to predict travel times along a specified sequence of links. Han et al. \cite{han2021multisemantic} addressed additional time delays at intersections by developing a multi-semantic path representation model, learning semantic representations for both link and intersection sequences, and combining non-Euclidean and Euclidean spatial information. Heterogeneous networks have also been used to address delay issues at intersections. Hong et al. \cite{hong2020heteta} introduced HetETA, which uses heterogeneous information graphs by converting the road map into a multi-relational network, capturing the directionality of intersections and applying temporal and graph convolutions to learn spatiotemporal representations. Some studies transform the TTE problem into a classification task. Ye et al. \cite{ye2022cateta} developed CatETA, categorizing travel times into discrete classes based on average time for each category, using BiGRU and embedding neural networks to extract spatiotemporal features. Liu et al. \cite{liu2023uncertainty} reframed travel time estimation as a multi-class classification problem to account for uncertainties, incorporating an adaptive local label-smoothing technique to capture ordinal relationships between travel time categories. While these route-based methods achieve high performance by utilizing rich features and advanced neural networks, they primarily focus on providing mean travel time estimates without accounting for the variations or confidence levels of their predictions.

\subsection{Spatiotemporal Probabilistic Regression}

The confidence level of estimation results is critical in addition to the predicted values. Probabilistic regression offers significant advantages by providing uncertainty estimates and enhancing the model's adaptability and reliability. Probabilistic methods grounded in statistical theory have gained attention in economic analysis \cite{engle2004caviar,gilboa2008probability}. Models like ARCH \cite{engle1982autoregressive} and GARCH \cite{bollerslev1986generalized} capture higher-order statistical moments, offering deeper insights into data behavior and providing more accurate guidance for economic decision-making. Recent developments have fused probabilistic regression with neural networks. Salinas et al. \cite{salinas2020deepar} introduced DeepAR, an RNN-based framework for probabilistic forecasting, offering confidence intervals for predictions and utilizing a negative log-likelihood function to accommodate variations within time series. Extending this work, they proposed a high-dimensional multivariate probabilistic forecasting method \cite{salinas2019high}, capturing joint distributions in multivariate time series by modeling covariance structures with a low-rank-plus-diagonal representation, optimizing parameter efficiency.

In travel time estimation, researchers have increasingly focused on probabilistic approaches to better characterize uncertainty and variability. Li et al. \cite{li2019learning} introduced a deep generative framework that models travel time distributions by leveraging real-time traffic conditions, integrating dynamic traffic patterns with static spatial features using an attention mechanism. Zhou et al. \cite{zhou2023travel} proposed a probabilistic method for learning temporal and spatial representations of road segments using an attributed graph, incorporating dynamic traffic states and road network structures, and designing a loss function based on negative log-likelihood. Mao et al. \cite{mao2023gmdnet} introduced GMDNet, combining graph neural networks with mixture density networks to predict travel time distributions in logistics systems, using the Expectation-Maximization algorithm for stable training and capturing complex dependencies. 

While these methods focus on modeling travel times for individual trips, they generally assume independence between trips. This neglects potential inter-trip correlations, leading to an incomplete representation of underlying dependencies and reducing the accuracy of both travel time predictions and uncertainty quantifications. In prior work \cite{xu2024link}, we developed a joint probability model for multi-trip scenarios, explicitly addressing correlations among trips. However, this approach encounters difficulties in handling time-varying dynamics and dealing with sparse, uneven coverage of trip data. Effectively modeling the temporal variability of multi-trip dependencies remains a significant challenge.

\section{Definitions and Problem Formulation} \label{sec:formulation}

\begin{figure*}
	\centering
	\includegraphics[width=7in]{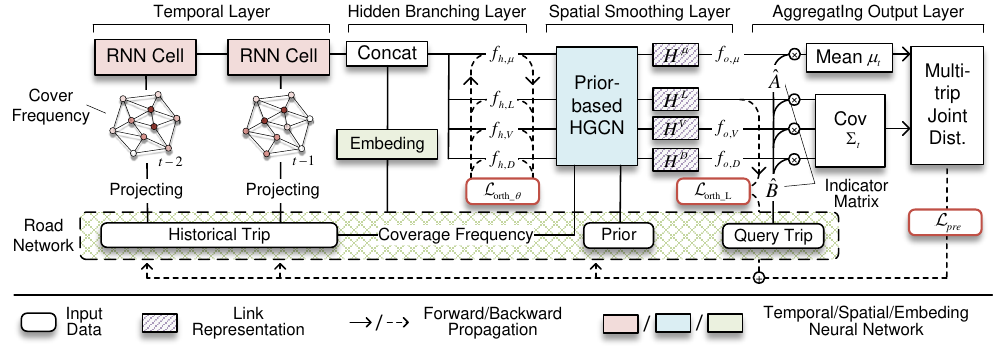}
	\caption{Overall architecture of \textit{SPTTE}.}
	\label{figure:overall}
\end{figure*}

\subsection{Definition: Road Network}

A road network consists of a complete set of road segments within a given area or city. It is represented by a graph $\mathcal{G} = (\mathcal{V}, \mathcal{E})$, where $\mathcal{V}$ is the set of nodes corresponding to road segments, and $\mathcal{E}$ is the set of edges capturing the topological relationships between these segments. In this paper, ``link'' and ``segment'' are used interchangeably to refer to the fundamental units of the road network.

\subsection{Definition: Links and Trips}

A link $l \in \mathcal{V}$ represents a uniquely indexed road segment within the road network. In this work, we follow OpenStreetMap (OSM) \cite{hakley2008open} for link labeling. A trip $T = \{l_1, ..., l_k\}$ is defined as a sequence of links, derived by mapping the vehicle's actual GPS trajectory onto the road network. The GPS information includes latitude, longitude, and timestamp $\xi $. The travel time of a trip is calculated as the difference between the timestamps of the last and first GPS points in the vehicle's trajectory, given by $\tau = \xi_k - \xi_1$.

\subsection{Problem Formulation} 

The core of our work is to model the evolution of the spatiotemporal stochastic process of trips based on fragmented observations from its realizations. Specifically, given a dataset $\mathcal{D} = \{T_i \mid i = 1, \dots, N\}$ containing $N$ historical trips, our objective is to capture the evolving probability distribution of the trip travel time, and then estimate the joint distribution of travel times for multiple query trips $\bm{\tau_q}$. This is formulated as
\begin{equation}
\bm{\tau_q} \sim \mathcal{N} \left(\bm{\mu_q}, \bm{\Sigma_q}\right)=\mathcal{F}_{dist.}(T_q;\bm{\theta_\mu},\bm{\theta_\Sigma}|\mathcal{D}),
\end{equation}
where the travel times of multiple trips are assumed to follow a joint Gaussian distribution, $\mathcal{F}_{dist.}(\cdot)$ is the model designed to estimate both the mean and covariance of travel time $\bm{\tau_q}$, $\bm{\theta}$ represents the model parameters trained on historical data $\mathcal{D}$.

\section{Methodology} \label{sec:method}

\subsection{Overview of \textit{SPTTE}}
This study proposes a spatiotemporal joint probability estimation model (\textit{SPTTE}) for predicting multi-trip travel times, as shown in Figure~\ref{figure:overall}. The model is designed to capture the temporal variability of the joint probability distribution from historical multi-trip travel times to estimate future distributions. Distinguishing itself from existing methods, our approach adopts a macroscopic perspective, placing greater emphasis on the stochastic process properties of the entire sample rather than merely concentrating on feature modeling for individual instances.

Specifically, we design a temporal Gaussian process parameterization to represent the travel time distribution over road network links. For each link, a hidden state is learned via gated recurrent units, capturing the temporal frequency characteristics of trips. This hidden state is then branched into four kinds of representations, which are used to generate the mean and covariance matrix of travel times. These representations undergo heterogeneity smoothing through a prior-based graph convolutional network to enhance the reliability of unevenly covered links. By leveraging two meticulously engineered trip indicator matrices and the contractility of Gaussian distributions, link embeddings are aggregated into low-rank trip representations, which are subsequently used to construct the joint distribution of travel times across multiple trips. The negative log-likelihood loss and two orthogonal constraints are used to optimize the link representations.

\subsection{Temporal Gaussian Process Parameterization for Link Travel Time Distribution}

Links can be regarded as the fundamental units of trips, with a finite set of links that can serve as a foundation for representing an infinite number of trips. In this section, we employ a Gaussian distribution to model the travel time for each link, accounting for the random variability introduced by different trips. Given that the historical data spans several days, we discretize each day into multiple time intervals (e.g., 20-minute slots). Here the discretization is utilized solely to define anchor points for capturing the periodic patterns in the travel time distribution. The travel time $t_{i,l,q}$ for time slot $i$, link $l$, and trip $q$ is formulated as
\begin{equation} \label{equ:decomp}
t_{i,l,q} = \mu_{i,l}  + \epsilon_{i,l,q},
\end{equation}
where $\mu_{i,l}$ represents the mean travel time for link $l$ during time slot $i$, and $\epsilon_{i,l,q}$ captures the trip-specific error, accounting for factors such as variability in driver behavior, heterogeneity in vehicle characteristics, and traffic incidents that propagate across multiple links.

We assume the trip-specific error $\bm{\epsilon}$ in Eq.~\eqref{equ:decomp} has a zero mean.
To model its covariance matrix $\bm{\Sigma}$ while ensuring it remains positive definite and conducive to optimization, we adopt a parameter-efficient approach utilizing a scale-separated low-rank-plus-diagonal parameterization:
\begin{equation} \label{equ:sigmap}
	\bm{\Sigma} = {\bm{V}^{\frac{1}{2}}} \bm{L}\bm{L}^{\top}{\bm{V}^{\frac{1}{2}}} + \bm{D},
\end{equation}
where $\bm{L}\in \mathbb{R}^{|\mathcal{V}|\times r_L}$ with $r_L\ll |\mathcal{V}|$ is a low-rank matrix, and $\bm{D}\in \mathbb{R}^{|\mathcal{V}|\times |\mathcal{V}|}$ is diagonal matrix with $\bm{D}_{l,l}>0$ for $l=1,\ldots,|\mathcal{V}|$. The matrix $\bm{V}\in \mathbb{R}^{|\mathcal{V}|\times |\mathcal{V}|}$ represents the covariance scale, which is also a diagonal matrix. The term $\bm{L}\bm{L}^{\top}$ represents the covariance matrix with the scale removed. Under this decomposition, $\bm{L}^{\top}$ and $\bm{L}$ are orthogonal to each other, allowing the introduction of an additional constraint, $\min(\bm{L}^{\top}\bm{L} - \bm{I})$, to facilitate optimization.
Thus, the joint distribution of travel times on all links on time slot $i$ can be modeled as
\begin{equation}\label{link-level}
	\bm{t}_i = \bm{\mu}_i+\bm{\epsilon}_i \sim \mathcal{N} \left(\bm{\mu}_i, \bm{\Sigma}_i\right),
\end{equation}
where $\bm{\mu}_i\in \mathbb{R}^{|\mathcal{V}|\times 1}$ and $\bm{\epsilon}_i \in \mathbb{R}^{|\mathcal{V}|\times 1}$ are vectorized mean and trip-specific error at time slot $i$, respectively. 

Furthermore, we consider that the link travel times at different time periods are autocorrelated $\Cov(\bm{t}_{i},\bm{t}_{i'}) \ne 0$, due to the smooth temporal variations in traffic conditions. This means that the distribution of future link travel times can be predicted from historical distributions.
We assume the existence of a latent probabilistic generative function ${\mathcal{H} }( \cdot )$ that generates the distribution of future travel times based on a limited history of past travel times. The conditional probability distribution of link travel time across time slots can be formulated as
\begin{equation}
	P({\bm{t}_i}|{\bm{t}_{i - 1}},...,{\bm{t}_{i - k}}) = {\mathcal{H} }({\bm{t}_{i - 1}},...,{\bm{t}_{i - k}};\bm{\theta} ).
\end{equation}
This can also be interpreted as the distribution of future link travel times being derived from a hidden state $\bm{h}_i$ generated by ${\mathcal{H}}(\cdot)$, which incorporates temporally dependent information of the historical distribution. We parametrize the temporal probability distribution using a Gaussian process:
\begin{equation}
	P({\bm{t}_i}|\bm{h}_i) = \mathcal{N}(\bm{t}_i|\mu({\bm{h}_i}),\Sigma({\bm{h}_i})).
\end{equation}

However, a significant obstacle is the inaccessibility of the distribution of historical link travel times $\bm{t}_i$, rendering it impractical to use as input for generating dynamic hidden states $\bm{h_i}$. To address this, we reflect the temporal variability of link travel time distribution based on an alternative covariate, the coverage frequency $\mathbb{F}$ of trips on the road network, for each period. It can be viewed as a random sampling of traffic volume in the form of a multivariate time series, reflecting the time-varying nature of traffic conditions. The coverage frequency $\mathbb{F}_{i,l}$ of time period $i$ and link $l$ is formulated as
\begin{equation}
	{\mathbb{F}_{i,l}} = \sum\limits_{j = 1}^{|{\mathcal{D}_i}|} {\sum\limits_{k = 0}^{|{T_j}|} {{\mathbb {I}}} }( {T_{j,k}} = {l}),
\end{equation}
where $|{\mathcal{D}_i}|$ represents the number of trips in the data subset ${\mathcal{D}_i}$ for time period $i$, $|{T_j}|$ represents the number of links in the trip $T_j$, $\mathbb {I}$ is a indicator function, such that ${\mathbb {I}} ( {T_{j,k}} = {l})=1$, and 0 otherwise.

The recurrent neural networks (e.g., Gated Recurrent Units) are utilized to capture the temporal dependencies of the alternative covariate $\mathbb{F}_{i-\eta:i-1}\in \mathbb{R}^{\eta \times |\mathcal{V}|}$ over the past $\eta$ time periods and model the hidden state $\bm{h}_i\in \mathbb{R}^{|\mathcal{V}|\times r_h}$. In addition to the dynamic hidden state $\bm{h}_i$ form GRU, a static link embedding vector $\bm{e}\in \mathbb{R}^{|\mathcal{V}|\times r_e}$ is concatenated to learn the global feature of link travel time. We employed four multilayer perceptrons $f_{h,s}(\cdot),s \in \{\mu,L,V,D\}$ as hidden layers to transform $\bm{x}_i=[\bm{h}_i,\bm{e}] \in \mathbb{R}^{|\mathcal{V}|\times (r_h+r_e)}$ into four branches in Eq.~\eqref{link-level}, and obtain link representation $\bm{H}^s_i$,
\begin{equation}\label{Hiddenstate}
\bm{H}^s_i=f_{h,s}(\bm{x}_i)=\bm{x}_i\bm{w}_h^s\in \mathbb{R}^{|\mathcal{V}|\times (r_h+r_e)},\ s \in \{\mu,L,V,D\}.
\end{equation}
where $\bm{w}^s_h\in \mathbb{R}^{ (r_h+r_e)\times (r_h+r_e)}$ are learnable parameters.
The hidden layer is used to differentiate specific features for different branches from the unified hidden state.
After passing through the output layer composed of multilayer perceptrons $f_{o,s}(\cdot),s \in \{\mu,V,D\}$, which perform regression to obtain the $\bm{\mu}$, $\bm{V}$, and $\bm{D}$ ($\bm{L}$ does not require regression), the four components of Eq.~(\ref{link-level}) can be expressed as follows:
\begin{equation}\label{componet1}
	{\bm{\mu} _{i}} = {f_{o,\mu} }({\bm{H}^\mu_{i}}) = {\bm{H}^\mu_{i}}{\bm{w}^\mu_o }\in \mathbb{R}^{|\mathcal{V}|\times 1},
\end{equation}
\begin{equation}\label{componet2}
	{\bm{L}_{i}} = {\bm{H}^L_{i}}\in \mathbb{R}^{|\mathcal{V}|\times (r_h+r_e)},
\end{equation}
\begin{equation}\label{componet3}
    \begin{aligned}
        {\bm{V}_{i}} &= \text{diag}(\log (1 + {\exp({{f_{o,V}}({\bm{H}^V_{i}})}})))\\
        &= \text{diag}(\log (1 + {\exp({{\bm{H}^V_{i}}{\bm{w}^V_o}}})))\in \mathbb{R}^{|\mathcal{V}|\times |\mathcal{V}|},
    \end{aligned}
\end{equation}
\begin{equation}\label{componet4}
    \begin{aligned}
    	{\bm{D}_{i}} &= \text{diag}(\log (1 + {\exp{({f_{o,D}}({\bm{H}^D_{i}}))}}))\\ &= \text{diag}(\log (1 + {\exp{({\bm{H}^D_{i}}{\bm{w}^D_o})}}))\in \mathbb{R}^{|\mathcal{V}|\times |\mathcal{V}|},
    \end{aligned}
\end{equation}
where we use a Softplus function to model the diagonal entries of $\bm{D}$ and $\bm{V}$, ensuring that the elements are positive, $\bm{w}^s_o \in \mathbb{R}^{ (r_h+r_e)\times 1}$ are learnable parameters. According to Eq.~\eqref{link-level}, the temporal parameterization of the link-level travel time distribution can be obtained.

\subsection{Prior-based Heterogeneity Smoothing Strategy for Link Representation}

The stability and reliability of the representation $\bm{H}^s_i$ in Eq.~\eqref{Hiddenstate} depend on a substantial amount of data; however, the uneven and sparse trip sample leads to coarse coverage of certain links, resulting in unreliable data-driven link representation learning. Although we employ the data augmentation techniques as in our previous work\cite{xu2024link}, challenges persist for links with very few or no trip coverage. To address this, we incorporate prior knowledge about the links, such as length and number of lanes, to heterogeneously correct the unreliable representations from the uneven observations.

Consider a neighborhood affected by traffic diffusion phenomena, where links with similar prior characteristics are likely to exhibit analogous posterior features\cite{saberi2020simple}. 
A prior similarity matrix $\bm{P}$ is defined based on prior knowledge to encapsulate the spatial dependency relationships among the links in a neighborhood. Given each link's prior feature vector \( \bm{c}_l \in \mathbb{R}^{1 \times h} \) (such as, length, number of lanes, etc.) and the adjacency matrix \( \bm{A}_{adj} \) of the road network, the prior similarity between nodes is formulated as
\begin{equation}\label{PMatrix}
\bm{P}_{l l'} = 
	\begin{cases} 
		\frac{1}{{1 + \sqrt{{ {\operatorname{dis}(\bm{c}_l - \bm{c}_{l'})}}}}},  & \text{if } (l, l') \in \mathcal{E} \\ 
            1, & l=l'\\
		0, & \text{otherwise} 
	\end{cases}.
\end{equation}
where the function $\operatorname{dis}(\cdot)$ is used to measure the distance between the prior feature $\bm{c}_l$ and $\bm{c}_{l'}$ (for example, in this case, we use the Euclidean distance), and the result is normalized to serve as the similarity weight between adjacent links.
For the self-similarity of each link, i.e., the diagonal elements of $\bm{P}$, self-loops are incorporated, with a value of 1 assigned to these elements.
For links that are not directly connected and do not exhibit explicit trip-level correlations, the prior similarity matrix is masked according to the adjacency relationships, assigning the value of 0 to the corresponding entries.

Therefore, the representations of nodes can be smoothed using the representations $\bm{H}_{\mathcal{N}(l)}$ from their neighboring nodes and the prior similarity weight  $\bm{P}$.
The smoothed representation can be expressed as
\begin{equation}\label{M1}
\bm{H}_l^{\text{new}} =\varphi_{hom} \cdot ( \bm{H}_{l}+ \sum_{l' \in \mathcal{N}(l)} \bm{P}_{l l'}  \bm{H}_{l'}),
\end{equation}
\begin{equation}
\varphi_{hom} =\frac{1}{{1 + \sum\limits_{l' \in \mathcal{N}(l)} {{\bm{P}_{ll'}}} }}.
\end{equation}
where $\varphi_{hom}$ is the normalization coefficient and we omit the time slot index $i$ for simplicity. The above is only a homogeneous smoothing based on prior similarity, which incorporates prior knowledge into the smoothing process, but it still cannot distinguish the reliability of the links.

\begin{figure}[!t]
	\centering
	\includegraphics[width=3.1in]{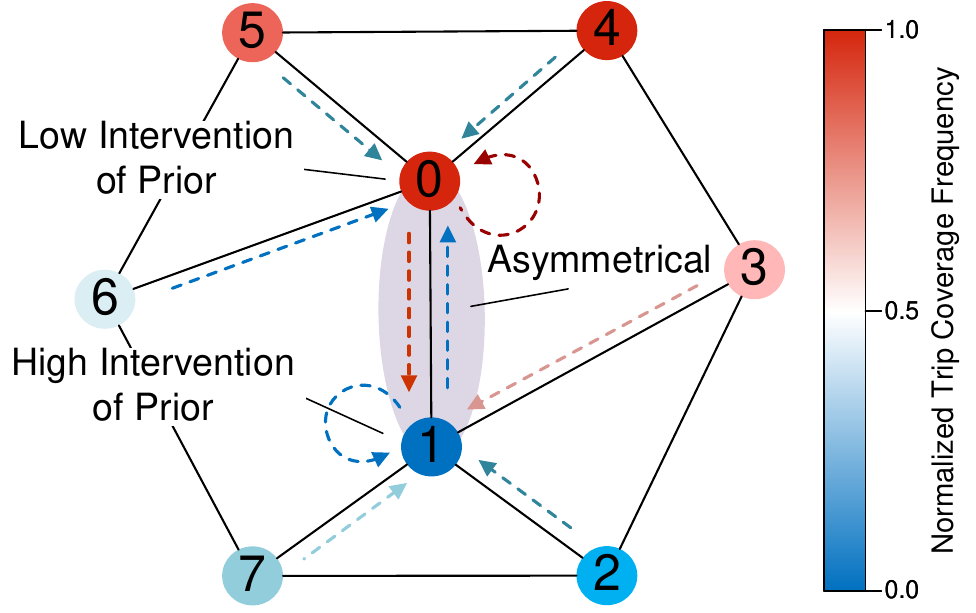}
	\caption{Process of Heterogeneity Smoothing. 
Introduce prior intervention based on heterogeneous trip coverage frequency to asymmetrically smooth link representations. Dashed lines represent smoothing weights, with red/blue indicating high/low weights. }
	\label{fig:smooth}
\end{figure}

The reliable link representations learned from large-scale data do not require intervention from prior knowledge. The superfluous incorporation of prior knowledge may undermine the features extracted through data-driven learning.
We consider that link representations learned under high trip coverage frequency are more reliable and design a frequency-based heterogeneous aware matrix $\bm{W}_f$ accordingly. Considering the saturation effect and gradual transition of confidence based on trip coverage\cite{vapnik2013nature}, we use an exponential decay function to define the weights of the diagonal elements,
\begin{equation}\label{Wfunction1}
\bm{W}_{f,ll} = 1 - \exp\left(-k_f \cdot \frac{{\mathbb{F}}_l}{{\max(\mathbb{F})}}\right),
\end{equation}
where $k_f$ is a learnable sensitivity factor. For the off-diagonal elements, higher weights are assigned to neighbors with greater frequencies to ensure the target link tends to obtain information from more reliable neighbors,
\begin{equation}\label{Wfunction2}
{\bm{W}_{f,ll'}} = \left\{ {\begin{array}{*{20}{c}}
		{\frac{{(1 - {\bm{W}_{f,ll}}){\mathbb{F}_{l'}}}}{{\sum\nolimits_{k \in {N_l}} {{\mathbb{F}_k}} }},\ \text{if}\ (l, l') \in \mathcal{E}}\\
		{\qquad0, \qquad \text{otherwise} \qquad}
\end{array}} \right..
\end{equation}
It is worth noting that $\bm{W}_f$ is an asymmetric matrix that facilitates the directed selection of reliable representations to smooth out unreliable ones. This strategy helps to prevent the dispersion of unreliable representations, as illustrated in Figure~\ref{fig:smooth}. In this context, links with high frequencies tend to prioritize their own representations, leading to $\bm{W}_{f,ll} \to 1$ and $\bm{W}_{f,ll'} \to 0$ for neighboring links. Conversely, links with low frequencies exhibit the opposite behavior, allowing for greater influence from their neighboring representation.

Therefore, heterogeneous smoothing is designed to handle link representations with varying levels of reliability differently. It is formulated as
\begin{equation}\label{M2}
\bm{H}_l^{\text{new}} =\varphi_{het} \cdot (\bm{W}_{f,ll} \bm{H}_{l}+ \sum_{l' \in \mathcal{N}(l)} \bm{W}_{f,ll'} \bm{P}_{l l'}  \bm{H}_{l'}),
\end{equation}
\begin{equation}
\varphi_{het}=\frac{1}{{\bm{W}_{f,ll} + \sum\limits_{l' \in \mathcal{N}(l)} {{\bm{W}_{f,ll'}\bm{P}_{ll'}}} }}.
\end{equation}
The message-passing mechanism of graph neural networks is used to realize the smoothing of representations, based on the prior similarity matrix $\bm{P}$ and the heterogeneous aware matrix $\bm{W}_f$. The Prior-based Heterogeneous Graph Convolutional Network (HGCN) is formulated as
\begin{equation}
{\bm{H}^{s,(k + 1)}} = \sigma (\boldsymbol{\tilde \Lambda }^s{\bm{H}^{s,(k)}}\bm{W}_g^{s,(k)}),s\in\{\mu, V, L,D\}
\end{equation}
where $\boldsymbol{\tilde \Lambda}^s=(\bm{D}^{s})^{-1} \bm{ \Lambda}^s$ is the normalized edge weight matrix, $\bm{D}^s$ is the degree matrix of $\bm{ \Lambda}^s$, $\bm{ \Lambda}^s=\bm{P}^s  \odot  \bm{W}_f$, $\odot$ represents the Hadamard product. $\bm{H}^{s,(k)}$ is the output of layer $k$ with the branch $s\in \{\mu,V,L,D\}$, $\bm{H}^{s,(0)}$ denotes the original link representation generated by the function $f_{h,s}(\cdot)$. Each branch can leverage specific prior knowledge to generate a prior similarity matrix $\bm{P}^s$. $\bm{W}_g^{s,(k)}$ is a trainable matrix of layer $k+1$ for branch $s$ and used to update the aggregated representation, and $\sigma$ represents the activation function. After this operation, the representations of unreliable links are smoothed based on prior similarity and their reliable neighbor links, while the reliable representations of links remain unaffected. The corrected link representation $\bm{H}^s$, after processing through Eq.~(\ref{componet1})–(\ref{componet4}), provides four more reliable components for the parameterization of the link travel time distribution.

\subsection{Joint Distribution Construction for Multi-trip Travel Time}

Trip samples are regarded as fragmented observations of realizations from the trip spatiotemporal stochastic process. Each trip sample only provides partial temporal and spatial information, as it cannot encompass the entire road network or time. In this section, we standardize the irregular and fragmented trip samples by projecting them onto the finite link representation space, thereby converting the complex spatiotemporal stochastic process of trips into a more structured link-based spatiotemporal stochastic process.

Specifically, we leverage the affine transformation properties of Gaussian distributions to construct trip representations from link space. Two indicator matrices are constructed based on the sequence of links traversed by the trip. Indicator matrix $\bm{A}=[a_{q,l}]\in \{0,1\}^{Q\times |\mathcal{V}|}$ is constructed for aggregating mean value ${\mu}$, where $Q$ is the number of trip samples, $a_{q,l}=1$ if trip $q$ covers link $l$ and 0 otherwise, and we denote the $q$-th row in $\bm{A}$ by $\bm{A}^q$. Indicator matrix $\bm{B}=\operatorname{blkdiag}(\{\bm{A}^q\})\in \mathbb{R}^{Q\times Q|\mathcal{V}|}$ as a block diagonal matrix composed of each row in $\bm{A}$ is constructed for model covariance of trips by
aggregating $\bm{L}, \bm{V},\bm{D}$. Here, we still adopt the data augmentation methods from our previous work \cite{xu2024link} to implement $k_{aug}$ subsampling augmentation on the trip sequences, and subsequently construct the corresponding augmented indicator matrices $\bm{\hat{A}}$ and $\bm{\hat{B}}$. 
Specifically, each row of the original indicator matrix $\bm{A}$ (where the elements are $1$) undergoes $k_{aug}$ equidistant subsampling, and the resulting subsamples are stacked to obtain the augmented matrix $\bm{\hat{A}}^q\in \{0,1\}^{(k_{aug}+1)\times |\mathcal{V}|}$, $\bm{\hat{A}}$ is composed by stacking $\{\bm{\hat{A}}^q\}_{q=1}^b$. $\bm{\hat{B}}=\operatorname{blkdiag}(\{\bm{\hat{A}}^q\})$ is used to construct the correlation of subsamples originating from the same trip. The joint distribution of multi-trip travel times $\bm{\tau}_{i}$ is derived as
\begin{equation} \label{equ:joint_aug}
	\bm{{\tau}}\sim \mathcal{N}\left(\bm{\hat{A}}\bm{\mu},  \bm{\hat{B}} (\bm{I}_{Q}\otimes \bm{\Sigma}) \bm{\hat{B}}^{\top}\right),
\end{equation}
where $\otimes$ represent the
Kronecker product operator, $b$ is batch size. In this way, the irregular multi-trip samples are represented in a regularized closed space with link representation serving as the base.

\subsection{Loss Function for Multi-trip Travel Time Estimation}
According to the distribution in Eq.~\eqref{equ:joint_aug}, the vectors $\bm{h}_i$, $\bm{e}$, and the projection functions ${f}_{h,s}(\cdot)$, ${f}_{o,s}(\cdot),s\in \{\mu,L,V,D\}$ require optimization. Leveraging historical trips, we adopt an empirical Bayes approach \cite{efron2012large} to estimate the parameters $\bm{\theta}$ in GRU and projection functions, by maximizing the marginal likelihood of all trip travel time observations. The posterior log-likelihood of observing trip travel time $\bm{ \tau}^*$ is
\begin{equation} \label{equ:loglik}
	\mathcal{L}_{pre}(\bm{ \tau}^*;\bm{\theta})\propto -\frac{1}{2}
	\left[\log \det (\bm{\tilde{\Sigma}})+\left(\bm{\tau}^*-\tilde{\bm{\mu}}\right)^{\top} \bm{\tilde{\Sigma}}^{-1}\left(\bm{\tau}^*-\tilde{\bm{\mu}}\right)\right],
\end{equation}
where $\bm{\tilde{\mu}}=\bm{\hat{A}}\bm{\mu}$ and $\bm{\tilde{\Sigma}}= \bm{\hat{B}} (\bm{I}_{Q}\otimes \bm{\Sigma}) \bm{\hat{B}}^{\top}$. From the computational perspective, the likelihood can be evaluated by applying the Woodbury matrix identity and the matrix determinant lemma. Specifically, $\bm{\tilde{\Sigma}}= \bm{\hat{B}} (\bm{I}_{Q}\otimes \bm{\Sigma}_p) \bm{\hat{B}}^{\top}$ takes the form of a block-diagonal matrix of dimension $b(k_{aug}+1)\times b(k_{aug}+1)$.  Both the inverse and the determinant of this matrix can be computed efficiently by performing operations on each of the $(k_{aug}+1)\times (k_{aug}+1)$ blocks individually.

Considering that the mean and covariance of the joint distribution are independent, but in the model, the vectors $\bm{H}^\mu,\bm{H}^L,\bm{H}^V,\bm{H}^D$, which are used to represent the mean and covariance, are derived from the same hidden state $\bm{h}_i$. This may result in model instability and diminished optimization efficiency. To address this, we introduce additional constraints to ensure that the parameters related to the expressions of the mean and covariance approach orthogonality, thereby enhancing their independence. It is formulated as
\begin{equation}
\mathcal{L}_{\text{orth}\_{\theta}}(\bm{\theta_\mu},\bm{\theta_L},\bm{\theta_V},\bm{\theta_D}) = \sum_{s \in \{L, V, D\}} \left( \frac{\bm{\theta}_\mu \cdot \bm{\theta}_s}{\|\bm{\theta}_\mu\| \|\bm{\theta}_s\|} \right)^2
\end{equation}
where $\bm{\theta}_s$ represent the parameters of the projection functions ${f}_{h,s}( \cdot ), s \in \{L,V,D\}$ in hidden branching layer. Furthermore, to ensure that the learned link representation vector $\bm{L}^\top$ is orthogonal to $\bm{L}$ in the covariance matrix as mentioned in Eq.~(\ref{equ:sigmap}), we introduce the loss function based on the Frobenius norm at the link level:
\begin{equation} \label{equ:loss_orth2}
	\mathcal{L}_{\text{orth}\_L}(\bm{\theta};\bm{L})=\|\bm{L}^\top \bm{L}-\bm{I}\|_F^2,
\end{equation}

Combining the three components of the loss function with appropriate weights, the final loss function is expressed as
\begin{equation}
	\begin{aligned}
		\mathcal{L}(\bm{\tau}^*;\bm{\theta}) &=  \mathcal{L}_{pre} + \alpha \mathcal{L}_{\text{orth}_\theta}+\beta \mathcal{L}_{\text{orth}\_L}\\
		&\propto - \frac{1}{2}
		\left[\log \det (\bm{\tilde{\Sigma}})+\left(\bm{\bm{\tau}}^*-\bm{\tilde{\mu}}\right)^{\top} \bm{\tilde{\Sigma}}^{-1}\left(\bm{\tau}^*-\tilde{\bm{\mu}}\right)\right] \\
		& + \alpha \sum_{s \in \{V, C, L\}} \left( \frac{\bm{\theta}_\mu \cdot \bm{\theta}_s}{\|\bm{\theta}_\mu\| \|\bm{\theta}_s\|} \right)^2+\beta \|\bm{L}^\top \bm{L}-\bm{I}\|_F^2,
	\end{aligned}
\end{equation}
where $\alpha$, $\beta$ are the weight coefficients used to trade off the learning rates between the two components of the loss function. The training process is summarized in Algorithms~\ref{alg:1}.

\begin{algorithm}[!t]
    \caption{Algorithm of SPTTE}
    \label{alg:1}
    \hspace*{0.02in} {\bf Input:} 
    \begin{itemize}
        \item Batches of trips $T=\{T_1,...T_b\}$,
        \item Road prior knowledge $\bm{A}_{adj},\bm{C^s},s \in \{\mu, V, L, D\}$
    \end{itemize}
    \hspace*{0.02in} {\bf Output:} 
    \begin{itemize}
        \item Travel time joint distribution $\bm{\tau} \sim \mathcal{N}(\bm{\mu},\bm{\Sigma})$
    \end{itemize}
    \vspace{0.1cm} 
    \hspace*{0.02in} {\bf Data Preprocessing:}
    \begin{algorithmic}[1]
        \State Initialize $\bm{A} \in \mathbb{R}^{b \times |\mathcal{V}|}$, $\bm{B} \in \mathbb{R}^{b \times (b \cdot |\mathcal{V}|)}$ with zeros,
        \For{$j=1$ to $b$}  
            \For{$l \in T_j$}  
                \State $\bm{A}[j, l] = 1$,\Comment{Indicator matrix $\bm{A}$}
                \State $\bm{B}[j, j \cdot |\mathcal{V}| + l] = 1$,\Comment{Indicator matrix $\bm{B}$}
            \EndFor
        \EndFor
        \State Stack to obtain the augmented indicator matrix $\bm{\hat{A}},\bm{\hat{B}}$,
        \State Statistic the trip coverage frequency $\mathbb{F}$,
        \State Calculate prior similarity matrix $\bm{P}$ based on Eq.~(\ref{PMatrix}),
        \State Calculate heterogeneous aware matrix $\bm{W}_f$ based on Eq.~(\ref{Wfunction1})(\ref{Wfunction2}),
    \end{algorithmic}
    \vspace{0.1cm} 
    \hspace*{0.02in} {\bf Training Process:}
    \begin{algorithmic}[1]
        \State $\bm{h}_i = GRU(\mathbb{F}_{i-\eta:i-1})$, \Comment{Temporal dependency capture}
        \State $\bm{e} = \text{Embedding}(\text{LinkIndex})$,\Comment{Global feature embedding}
        \State $\bm{H}_i^s = f_{h,s}([\bm{h}_i, \bm{e}])$,\Comment{Original representation}
        \State $\boldsymbol{\tilde \Lambda}^s=(\bm{D}^{s})^{-1} \bm{P}^s  \odot  \bm{W}_f$,\Comment{Edge weight matrix}
        \State $\bm{H}_i^{s,new} = \sigma (\boldsymbol{\tilde \Lambda }^s{\bm{H}^{s}_i}\bm{W}_g^{s})$,\Comment{Smoothed representation}
        \State $\bm{\mu} = \bm{\hat{A}} f_{o, \mu}(\bm{H}_i^{\mu, new})$,\Comment{Mean}
        \State $\bm{L} = \bm{H}_i^{L, new}$,
        \State $\bm{V} = \text{diag}(\log(1 + \exp(f_{o, V}(\bm{H}_i^{V, new}))))$,
        \State $\bm{D} = \text{diag}(\log(1 + \exp(f_{o, D}(\bm{H}_i^{D, new}))))$,
        \State $\bm{\Sigma} = \bm{\hat{B}} \left( \bm{I}_b \otimes \left( \bm{V}^{-\frac{1}{2}} \bm{L} \bm{L}^{\top} \bm{V}^{-\frac{1}{2}} + \bm{D} \right) \right) \bm{\hat{B}}^\top$,\Comment{Cov}
        \State Compute loss $\mathcal{L}(\bm{\tau}^*;\bm{\theta})= \mathcal{L}_{pre} + \alpha \mathcal{L}_{\text{orth}\_\theta}+\beta \mathcal{L}_{\text{orth}\_L}$,
        \State Backpropagation and update model parameters
    \end{algorithmic}
\end{algorithm}

\section{Experiment} \label{sec:results}

\subsection{Datasets and Baselines}
We evaluate \textit{SPTTE} on two publicly available GPS trajectory datasets from ride-hailing services. \begin{itemize}
    \item \textbf{Chengdu}: This dataset includes 3,186 links and 346,074 trip samples collected over 6 days. Travel times range from 420 to 2,880 seconds, with a mean of 786 seconds. 
    \item \textbf{Harbin}: This dataset contains 8,497 links and 1,268,139 trip samples over 6 days. Travel times range from 420 to 2,994 seconds, with a mean of 912 seconds. 
\end{itemize}

We compare SPTTE against six state-of-the-art models, including three deterministic and three probabilistic methods.

Deterministic models:
\begin{itemize}
\item DeepTTE\cite{wang2018when}: It is designed to capture spatial and temporal dependencies directly from raw GPS sequences by employing a geo-based convolutional layer alongside recurrent neural networks.
\item HierETA\cite{chen2022interpreting}: It adopts a hierarchical approach to estimate the time of arrival, utilizing segment-view, link-view, and intersection representations to capture local traffic conditions, common trajectory patterns, and indirect influences, respectively.
\item MulT-TTE\cite{liao2024multifaceted}: It employs a multi-perspective route representation framework that integrates trajectory, attribute, and semantic sequences, alongside a path-based module and a self-supervised learning task, to improve context awareness and segment representation for travel time estimation.
\end{itemize}

Probabilistic models:
\begin{itemize}
\item DeepGTT\cite{li2019learning}: It leverages a convolutional neural network to represent real-time traffic conditions, and incorporates spatial embeddings, and amortized road segment modeling to model travel time distributions.
\item GMDNet\cite{mao2023gmdnet}: It captures spatial correlations utilizing a graph-cooperative route encoding layer, and uses a mixture density decoding layer for travel time distribution estimation.
\item ProbETA \cite{xu2024link}: It is a hierarchical model for joint probability estimation of multi-trip travel times. It discretely learns both inter-trip and intra-trip correlations of travel times, generates two sets of link representations, and calculates the conditional probability distribution for the query trip’s travel time.
\end{itemize}
The code and data are available at \url{https://github.com/ChenXu02/SPTTE}.

We divide each dataset into training, validation, and testing sets, allocating 70\% for training, 15\% for validation, and 15\% for testing. For the configuration of the \textit{SPTTE} model, we set a batch size of $256$ and perform trip augmentation with $k_{\text{aug}} = 5$. The embedding dimensions for link representations are set to $32$ for $r_h$ and $32$ for $r_e$. A time interval of 20 minutes is used, with GRU input sequence covering $\eta =6$ consecutive periods (a total of 2 hours). The GRU consists of 2 layers, each with a hidden layer dimension of $256$. The HGCN consists of 1 layer. The multi-layer perceptrons $f_{h,s}(\cdot)$ and $f_{o,s}(\cdot)$ have dimensions of $128$. The loss function includes trade-off coefficients $\lambda$ and $\beta$, both set to $0.02$. Link length is incorporated as prior knowledge of $\bm{H}^\mu, \bm{H}^V, \bm{H}^D$, and the unmodified adjacency matrix serves as prior knowledge for $\bm{H}^L$, as it is the representation after the scale is stripped.

Each model reaches its best performance after training for 100 epochs, utilizing a 12th Gen Intel(R) Core(TM) i9-12900K CPU alongside an NVIDIA Tesla V100 GPU. 
We employ four standard evaluation metrics to assess performance: Root Mean Square Error (RMSE), Mean Absolute Error (MAE), Mean Absolute Percentage Error (MAPE), and Continuous Ranked Probability Score (CRPS):
\begin{align}
&\text{RMSE} = \sqrt {\frac{1}{{\left| {{{\rm{\mathcal{D}}}}}  \right|}}\sum\nolimits_{j \in {{{\rm{\mathcal{D}}}}} } {{{({{\tilde \tau_{j}} - { \tau^*_{j}}})}^2}} },
\end{align}
\begin{align}
&\text{MAE} = \frac{1}{{\left| {{{\rm{\mathcal{D}}}}}  \right|}}\sum\nolimits_{j \in {{{\rm{\mathcal{D}}}}} } {\left| {{\tilde \tau_{j}} - { \tau^*_{j}}} \right|},
\end{align}
\begin{align}
&\text{MAPE} = \frac{1}{{\left| {{{\rm{\mathcal{D}}}}}  \right|}}\sum\nolimits_{j \in {{{\rm{\mathcal{D}}}}} } {\left| {\frac{{{{\tilde \tau_{j}} - { \tau^*_{j}}}}}{{{{ \tau^*_{j}}}}}} \right|},
\end{align}
\begin{align}
&\text{CRPS} = \frac{1}{{\left| \mathcal{D} \right|}}\sum\nolimits_{j \in D} {\int {{{(F({\tau _j}) - {1_{\{ {\tau _j} \ge {{ \tau }^*_j}\} }})}^2}d{\tau _j}} } ,
\end{align}
where $F(\tau_j)$ denote the cumulative distribution function of trip travel time $\tau_j$. $\tilde \tau_j$ represents a randomly sampled value from $\tau_j$ and $ \tau^*_j$ is the observed value.

\begin{table*}
	\centering
	\caption{Model performance comparison on Chengdu and Harbin datasets.}
	\label{tab:2}
	\begin{tabular}{c c c c c c c c c c} 
		\hline
		\multirow{2}{*}{Model}&  \multicolumn{4}{c}{Chengdu}& &\multicolumn{4}{c}{Harbin}  \\
		&  RMSE & MAE & MAPE(\%) & CPRS & & RMSE & MAE & MAPE(\%) & CPRS \\
		\hline
		{DeepTTE} & {$180.96$} & {$130.02$} & {$17.15$} & {---} &  & {$224.31$} & {$162.52$} & {$18.31$} & {---} \\
		{HierETA} & {$155.17$} & {$111.29$} & {$14.69$} & {---} &  & {$187.95$} & {$136.41$} & {$15.60$} & {---} \\
		{MulT-TTE} & \underline{$149.92$} & \underline{$105.22$} & \underline{$13.69$} & {---} &  & \underline{$178.33$} & {$129.77$} & \underline{$14.83$} & {---} \\
		{DeepGTT} & {$165.15$} & {$118.68$} & {$15.64$} & {$1.46$} &  & {$199.25$} & {$143.95$} & {$16.40$} & {$1.56$} \\
		{GMDNet} & {$151.50$} & {$107.75$} & {$13.96$} & \underline{$1.30$} &  & \underline{$176.78$} & \underline{$127.81$} & \underline{$14.49$} & \underline{$1.41$} \\
		{ProbETA} & {${138.36}$} & {${97.45}$} & {${12.54}$} & {${1.19}$} &  & {${159.68}$} & {${115.79}$} & {${13.17}$} & {${1.29}$} \\
		\textbf{\textit{SPTTE}} & {$\textbf{134.79}$} & {$\textbf{94.81}$} & {$\textbf{12.25}$} & {$\textbf{1.17}$} &  & {$\textbf{157.60}$} & {$\textbf{113.62}$} & {$\textbf{12.96}$} & {$\textbf{1.26}$} \\
		\hline	
		{Improvement} & {${10.09\%}$} & {${9.89\%}$} & {${10.53\%}$} & {${10.00\%}$} &  & {${10.85\%}$} & {${11.10\%}$} & {${10.57\%}$} & {${10.64\%}$} \\
		\hline	
	\end{tabular}
\end{table*}

Table~\ref{tab:2} presents the experimental results, showing that the proposed \textit{SPTTE} model consistently outperforms all baseline models. On the Chengdu dataset, \textit{SPTTE} surpasses deterministic baselines by over 1.44\% in MAPE and exceeds probabilistic baselines by more than 1.71\% in MAPE and 0.13 in CRPS, achieving an average relative improvement of 10.13\% against the strongest baseline. Similarly, on the Harbin dataset, \textit{SPTTE} outperforms deterministic baselines by over 1.87\% in MAPE and probabilistic baselines by more than 1.53\% in MAPE and 0.15 in CRPS, leading to an average relative improvement of 10.79\%. Notably, compared to our previous model, ProbETA, the \textit{SPTTE} framework demonstrates further advancements in both performance and efficiency due to its more generalized structure and reduced parameter complexity. Specifically, \textit{SPTTE} achieves reductions of 0.29\% (a relative reduction of 2.31\%) in MAPE and 0.02 (a relative reduction of 1.68\%) in CRPS on the Chengdu dataset, and reductions of 0.21\% (a relative reduction of 1.59\%) in MAPE and 0.03 (a relative reduction of 2.33\%) in CRPS on the Harbin dataset.

\begin{figure}
	\setlength{\abovecaptionskip}{-0cm}
	\centering
	\subfigure[]{\includegraphics[width=0.24\textwidth]{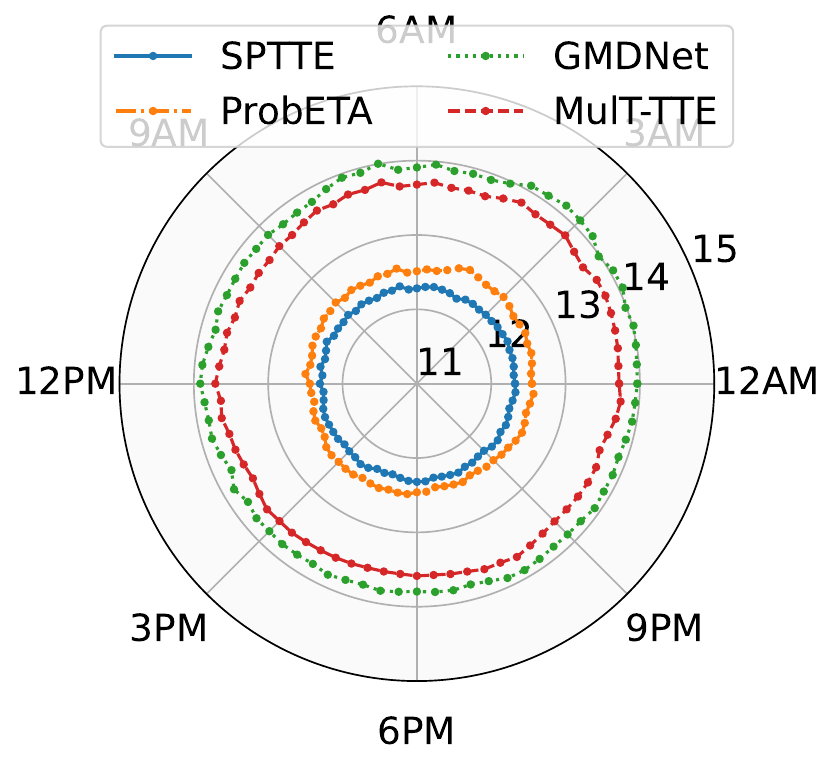}} 
	\subfigure[]{\includegraphics[width=0.24\textwidth]{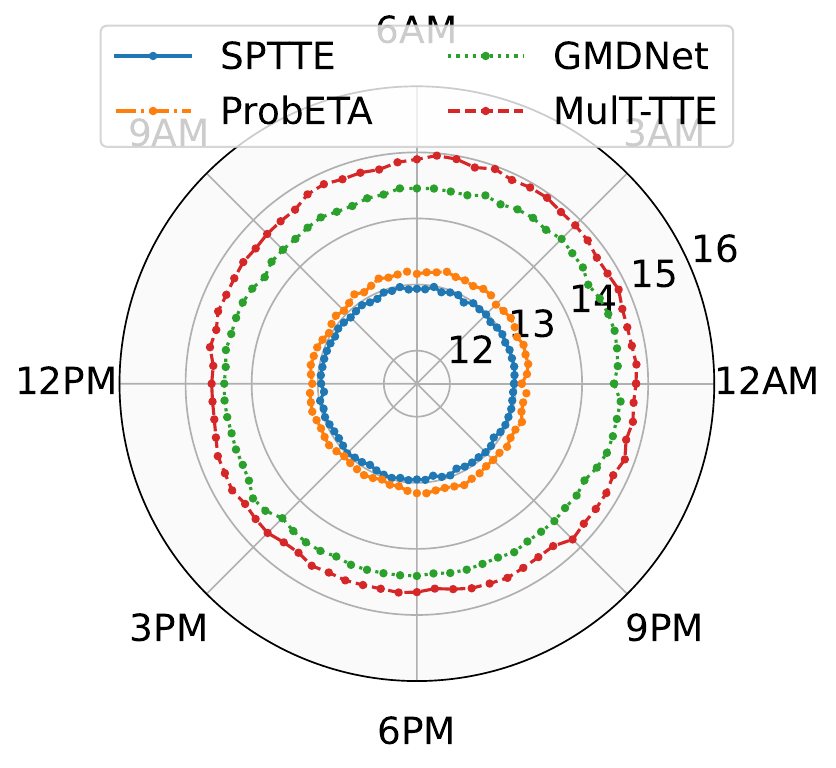}} 
	\caption{Estimation error by time period in a day. (a). MAPE in Chengdu dataset. (b). MAPE in Harbin dataset.}
	\label{fig:meanpie}
\end{figure}

We further visualized the estimation errors (MAPE) of our proposed model, \textit{SPTTE}, alongside those of ProbETA, the best probabilistic baseline (GMDNet), and the best deterministic baseline (MulT-TTE), across various time slots throughout the day, as depicted in Figure~\ref{fig:meanpie}. The predictive performance of all models fluctuates across different periods, likely due to the varying degrees of model optimization caused by the sparse and uneven distribution of trip samples. During peak hours, the sample size is significantly larger, leading the models to focus more on learning these dominant data patterns. In contrast, the smaller sample size during nighttime results in weaker features, which tend to be overshadowed during training, ultimately degrading the models' performance. Nonetheless, our model demonstrates more stable performance across all time periods. The evolution modeling of trip travel time distributions effectively captures the temporal dynamics inherent in the data. Through this learned evolutionary relationship, the training performance during high-sample periods positively influences the prediction accuracy in low-sample periods, mitigating the performance disparities caused by the uneven distribution of samples.

Additionally, we compare the performance of \textit{SPTTE} against ProbETA by reducing the size of the dataset along temporal and spatial dimensions. For the temporal dimension, we randomly remove 20\% and 40\% of the data from each time period. For the spatial dimension, we randomly select 10\% and 20\% of the links and remove all trips that pass through these links, while keeping the trips used in the test phase unchanged. This approach constructs scenarios with temporal sparse and spatial uneven sample distributions, where query trips involve links that have no historical trips passing through them. The experimental results are presented in Table~\ref{tab:3}.

\begin{table*}
	\centering
	\caption{Model performance comparison on Chengdu and Harbin datasets with temporal/spatial sparse.}
	\label{tab:3}
	\renewcommand {\arraystretch}{1.3}
	\resizebox{\textwidth}{19mm}
    {
	\begin{tabular}{c c c c c c c c c c c} 
		\hline
		\multirow{2}{*}{Sparsity}&\multirow{2}{*}{Model}&  \multicolumn{4}{c}{Chengdu}& &\multicolumn{4}{c}{Harbin}  \\
		\cline{3-6}\cline{8-11}
		 (temporal/spatial)& &  RMSE & MAE &MAPE(\%)& CPRS & &RMSE & MAE &MAPE(\%)& CPRS\\
		\hline
		\multirow{2}{*}{100\%}&{ProbETA} & {${138.36}$} &{${97.45}$}& {${12.54}$}&{${1.19}$}& {$ $}  & {${159.68}$}&{${115.79}$}& {${13.17}$}&{${1.29}$}  \\
		&{\textit{SPTTE}}& {${134.79}$} &{${94.81}$}& {${12.25}$}&{${1.17}$}& {$ $}  & {${157.60}$}&{${113.62}$}& {${12.96}$}&{${1.26}$}  \\
		\hline	
		\multirow{2}{*}{80\% / 90\%}&{ProbETA} & {${144.14/147.82}$} &{${101.56/104.57}$}& {${13.07/13.45}$}&{${1.24/1.28}$}& {$ $}  & {${165.65/170.16}$}&{${120.91/124.14}$}& {${13.74/14.11}$}&{${1.35/1.38}$}  \\
		&{\textit{SPTTE}}& {${138.07/140.60}$} &{${96.90/99.14}$}& {${12.54/12.79}$}&{${1.19/1.21}$}& {$ $}  & {${161.55/164.63}$}&{${117.91/119.86}$}& {${13.40/13.62}$}&{${1.31/1.33}$}  \\
		\hline	
		\multirow{2}{*}{60\% / 80\%}&{ProbETA} & {${151.44/160.34}$} &{${106.10/113.32}$}& {${13.75/14.62}$}&{${1.30/1.39}$}& {$ $}  & {${175.15/184.51}$}&{${128.11/134.62}$}& {${14.51/15.30}$}&{${1.42/1.50}$}  \\
		&{\textit{SPTTE}}& {${141.91/148.41}$} &{${100.72/104.56}$}& {${12.98/13.49}$}&{${1.23/1.28}$}& {$ $}  & {${166.75/174.11}$}&{${121.53/126.90}$}& {${13.82/14.43}$}&{${1.35/1.41}$}  \\
		\hline	
	\end{tabular}}
\end{table*}

The results demonstrate that \textit{SPTTE} outperforms ProbETA in both the temporal and spatial dimensions under sparse data conditions. In the temporal dimension, applying uniform sparsification (80\%, 60\%) across all links has a relatively small impact on the performance of both models, with \textit{SPTTE} exhibiting an advantage in predictive performance. It achieves an average reduction of 0.58\% (a relative reduction of 4.22\%) in MAPE and a decrease of 0.058 (a relative reduction of 4.33\%) in CRPS. In the spatial dimension, data sparsity (90\%, 80\%) significantly affects model performance; however, \textit{SPTTE} effectively utilizes prior knowledge as a foundation, smoothing the representations of under-trained links based on those of neighboring links. Compared to ProbETA, \textit{SPTTE} achieves an average reduction of 0.79\% (a relative reduction of 5.44\%) in MAPE and a decrease of 0.080 (a relative reduction of 5.75\%) in CRPS. Therefore, \textit{SPTTE} demonstrates greater robustness and improved predictive performance, exhibiting excellent adaptability and extrapolation capabilities for links with no or limited training under uneven sample distributions.

\begin{table}
\small
\centering
\caption{Ablation experiment on Chengdu/Harbin dataset.}
\label{tab:4}
\renewcommand {\arraystretch}{1.3}
{
	\resizebox{89mm}{11mm}{
\begin{tabular}{c c c c c } 
\hline
{Model}& {w/o SS}  & {w/o PK}  & {w/o HW}&\textit{SPTTE}  \\
\hline
RMSE& $149.90/174.85$ & $144.93/169.88$  &$147.77/172.63$&$138.52/168.18$  \\

MAE& $105.67/127.49$ & $101.40/123.77$  &$104.12/126.01$&$100.97/123.83$    \\

MAPE& $13.63/14.50$ & $13.15/14.07$  &$13.46/14.33$&$13.02/13.96$  \\

CRPS& $1.29/1.42$ & $1.25/1.38$  &$1.28/1.40$&{${1.24/1.37}$} \\
\hline	
\end{tabular}}}
\end{table}

\subsection{Ablation Study of Spatial Smoothing}

To further validate the effectiveness of the spatial smoothing layer and its components, we constructed three variants of \textit{SPTTE} focusing on the whole Spatial Smoothing, Prior Knowledge and Heterogeneous Weights:
\begin{itemize}

\item \textit{SPTTE}-w/o SS: All spatial smoothing operations are removed, and the hidden branching layer is directly connected to the aggregating output layer.

\item \textit{SPTTE}-w/o PK: The prior knowledge is removed, and the model determines the similarity of links based on the adjacency relationships of the road network.

\item \textit{SPTTE}-w/o HW: The heterogeneous weights are removed, and the model homogeneously introduces prior knowledge for all links to perform smoothing.
\end{itemize}
We conducted ablation experiments on the Chengdu and Harbin datasets with 20\% temporal sparsity and 10\% spatial sparsity to analyze the effects of those components.

\begin{figure}[!t]
	\centering
	\includegraphics[width=3in]{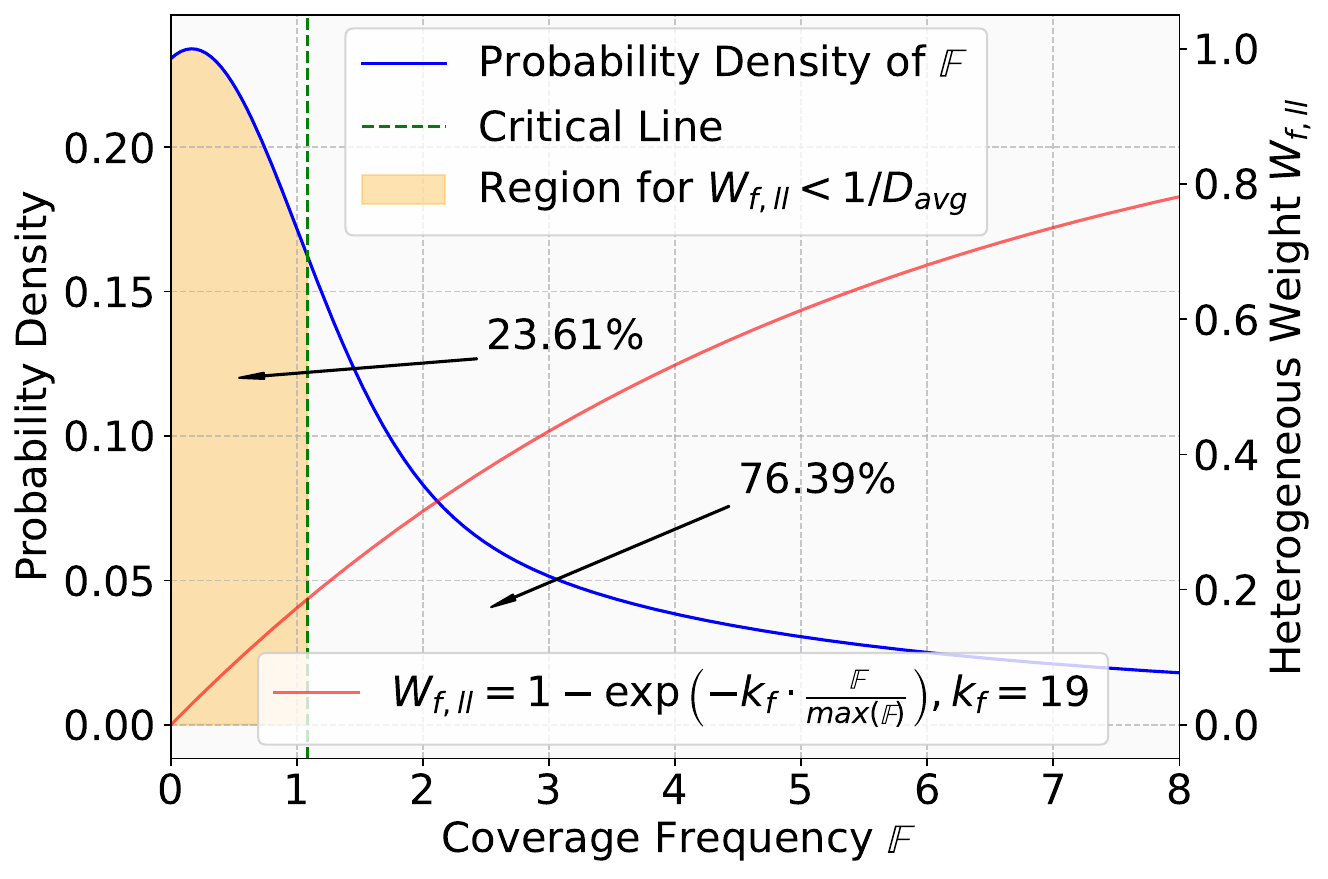}
	\caption{Coverage frequency distribution and heterogeneous weight function.}
	\label{fig:cont}
\end{figure}

The results of the ablation experiments are shown in Table~\ref{tab:4}. We can see that spatial smoothing plays a significant role in enhancing predictive performance and quantifying uncertainty under sparse data conditions. In this case, the entire spatial smoothing layer reduced MAPE by an average of 0.58\% (a relative reduction of 4.10\%) and lowered CRPS by 0.050 (a relative reduction of 3.70\%). Among these, the heterogeneity weights played a major role, achieving an average reduction of 0.41\% (a relative reduction of 2.93\%) in MAPE and 0.035 (a relative reduction of 2.63\%) in CRPS. The incorporation of prior knowledge provided a positive enhancement of 0.12\% (a relative reduction of 0.89\%) in MAPE and 0.01 (a relative reduction of 0.4\%) in CRPS. 

We visualized the learned heterogeneous weight construction function Eq.~\ref{Wfunction1} and compared it with the distribution of coverage frequency $\mathbb{F}$, as shown in Figure~\ref{fig:cont}. Based on the average degree of the graph (including self-loops) $D_{avg}=5.37$, we define $\bm{W}_{f,ll}=1/D_{avg}$ as the critical value on the figure, which means it is the average weight that each node should receive under homogeneous conditions. When $\bm{W}_{f,ll}$ is below this critical value, it indicates that the reliability of the node is low, and it tends to absorb information from its neighbors. Conversely, when $\bm{W}_{f,ll}$ is over this critical value, it is inclined to disseminate its information to neighboring nodes. From the graph, it can be observed that the critical coverage frequency is 1.08. 76.39\% of the nodes are relatively reliable, while 23.61\% are less reliable, requiring more correction based on prior knowledge and neighboring information.

\begin{figure}
	\setlength{\abovecaptionskip}{-0cm}
	\centering
	\subfigure[]{\includegraphics[width=0.24\textwidth]{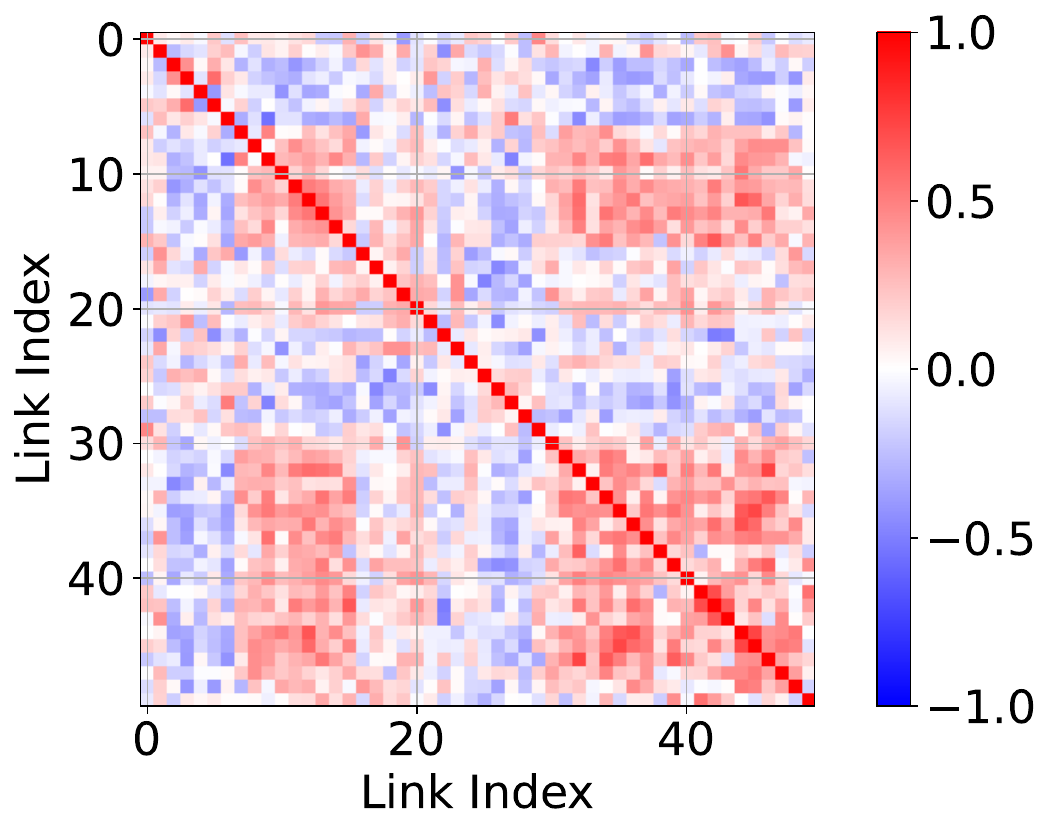}} 
	\subfigure[]{\includegraphics[width=0.24\textwidth]{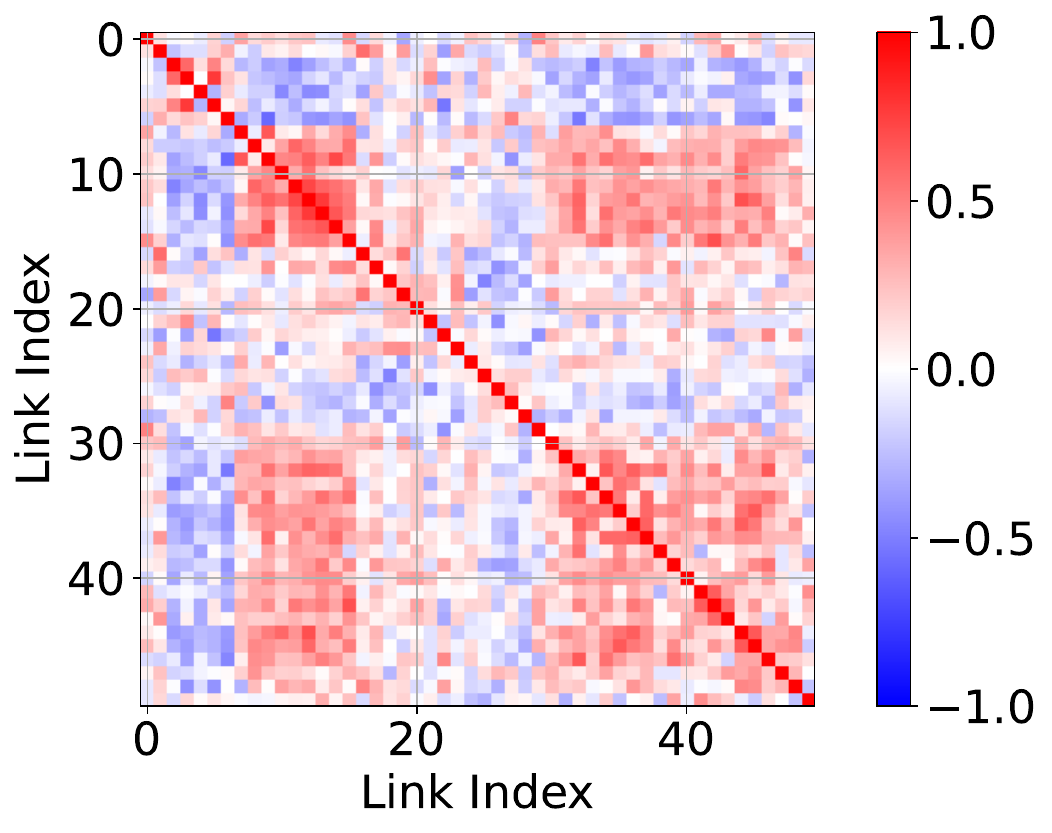}} 
	\subfigure[]{\includegraphics[width=0.24\textwidth]{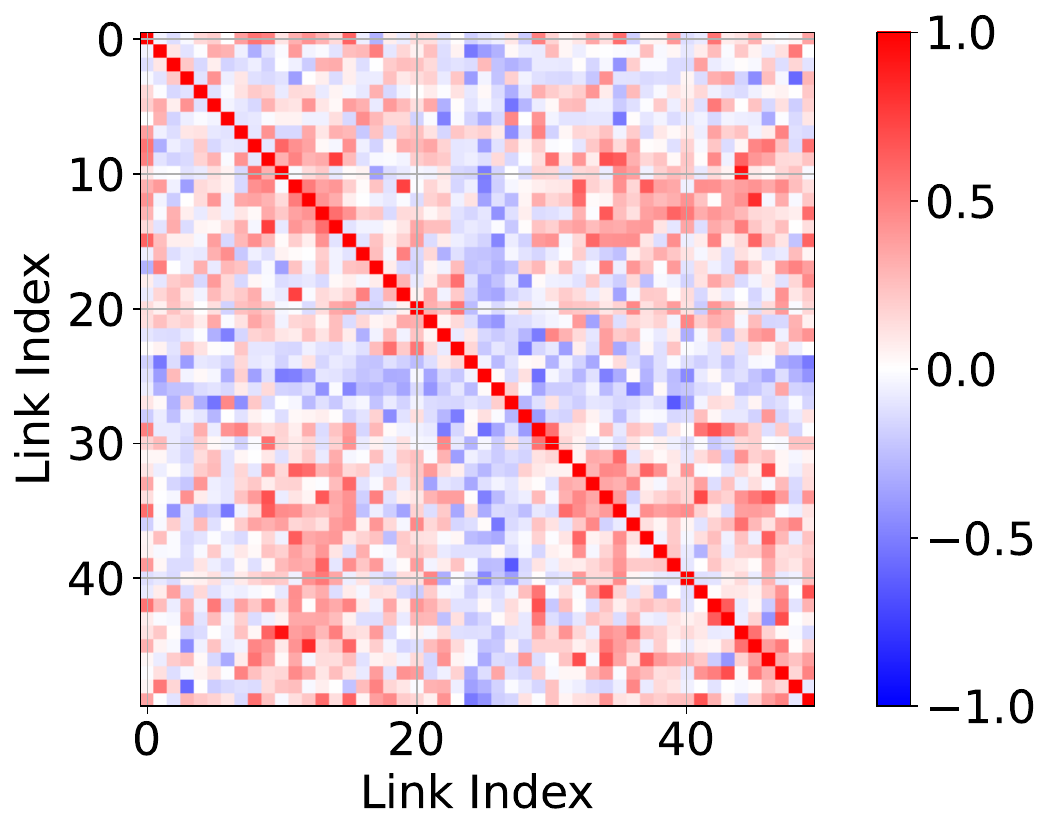}} 
	\subfigure[]{\includegraphics[width=0.24\textwidth]{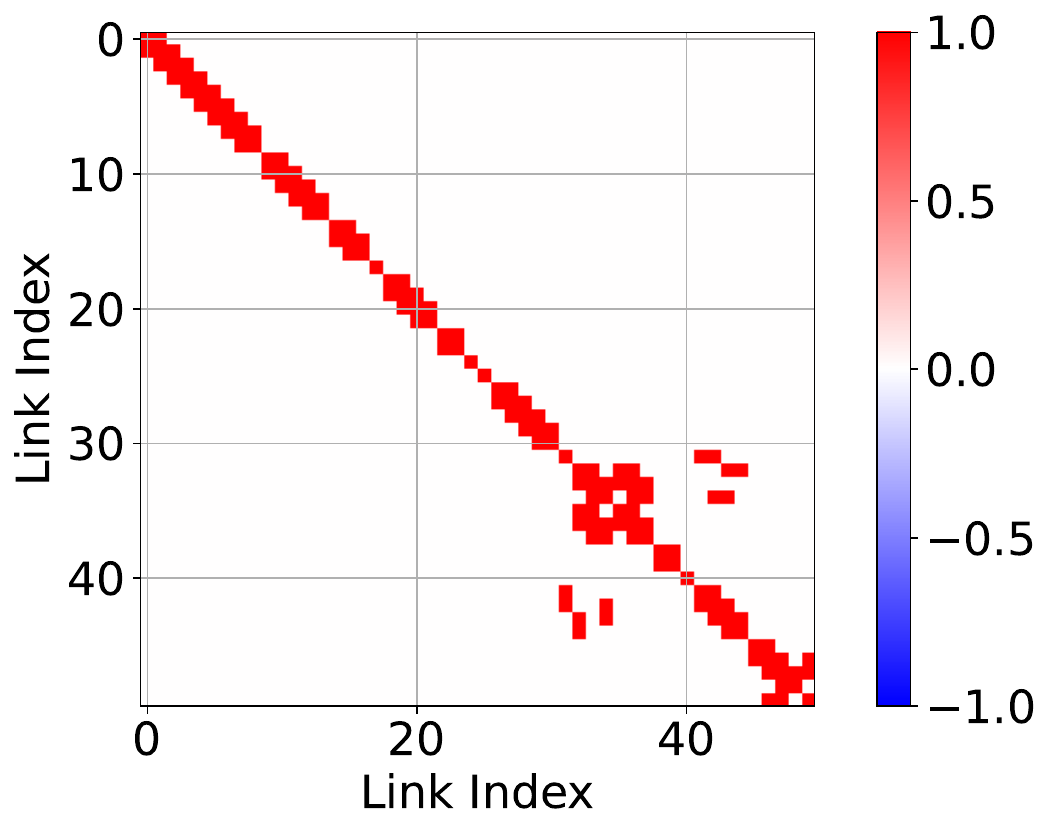}} 
	\caption{Visualization of learned link correlation. (a). Link correlation during morning peak hours (9:00-9:20 AM). (b). Link correlation during evening peak hours (5:00-5:20 PM). (c). Link correlation during off-peak hours (3:00-3:20 AM). (d). The real adjacency matrix of the road network.}\label{fig:cor}
\end{figure}

\subsection{Spatiotemporal Interpretability of Link Representation}

We analyze the spatiotemporal interpretability of link embedding vectors on the Harbin dataset. Firstly, we compared the learned correlation matrices from \textit{SPTTE} with the real adjacency matrix, focusing on visualizing the link correlations during morning peak hours (9:00-9:20 AM), evening peak hours (5:00-5:20 PM) and off-peak hours (3:00-3:20 AM). As shown in Figure~\ref{fig:cor}, during morning and evening peak hours, links exhibit stronger and more diverse correlations, with slight differences in the two distributions. Additionally, there is a significant similarity to the distribution of the true adjacency matrix. In contrast, link correlations during off-peak hours are weaker and show minimal differences between links, but they still reflect some characteristics of link adjacency. We believe that the high traffic volumes (congestion) during peak periods make trip-specific features more pronounced across links. Conversely, during off-peak hours, the sparse vehicle distribution results in the state of free flow on most links, leading to less pronounced differences in link correlations.

\begin{figure}
	\setlength{\abovecaptionskip}{-0cm}
	\centering
	\subfigure[]{\includegraphics[width=0.24\textwidth]{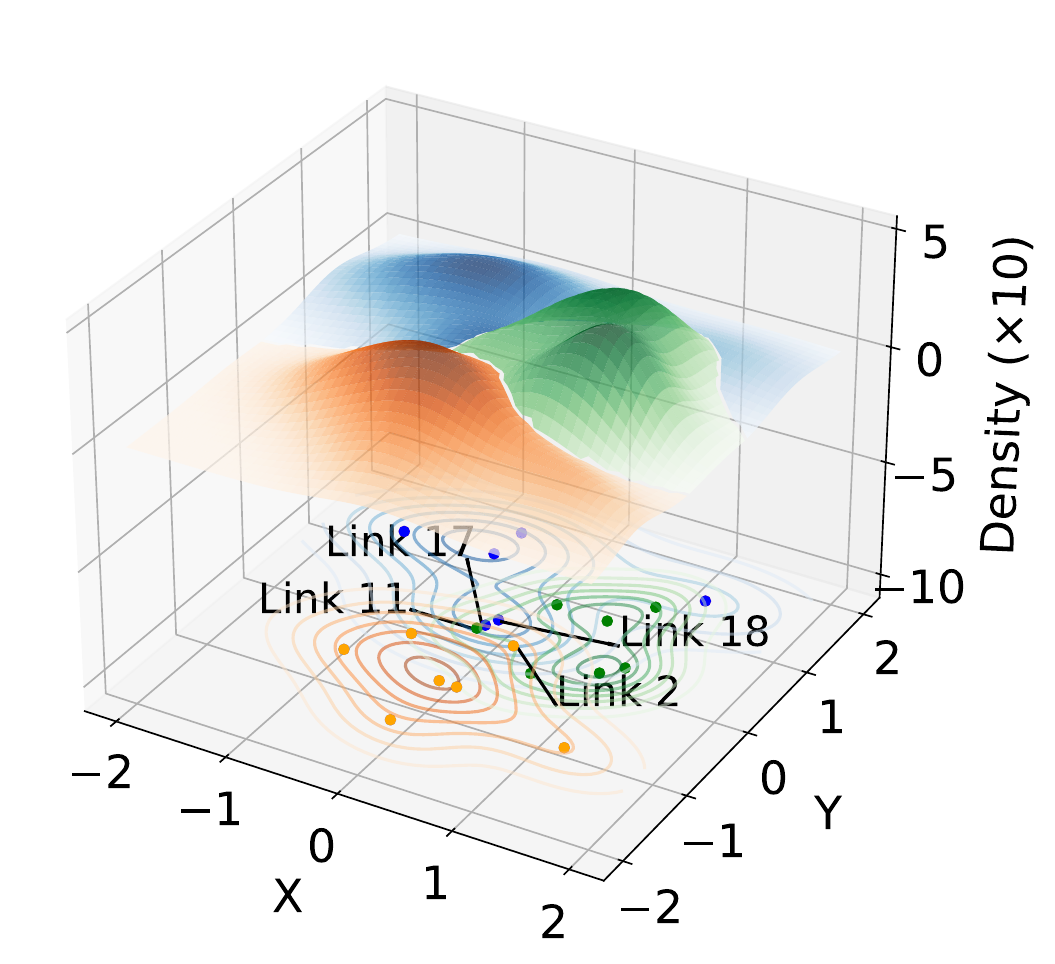}} 
	\subfigure[]{\includegraphics[width=0.24\textwidth]{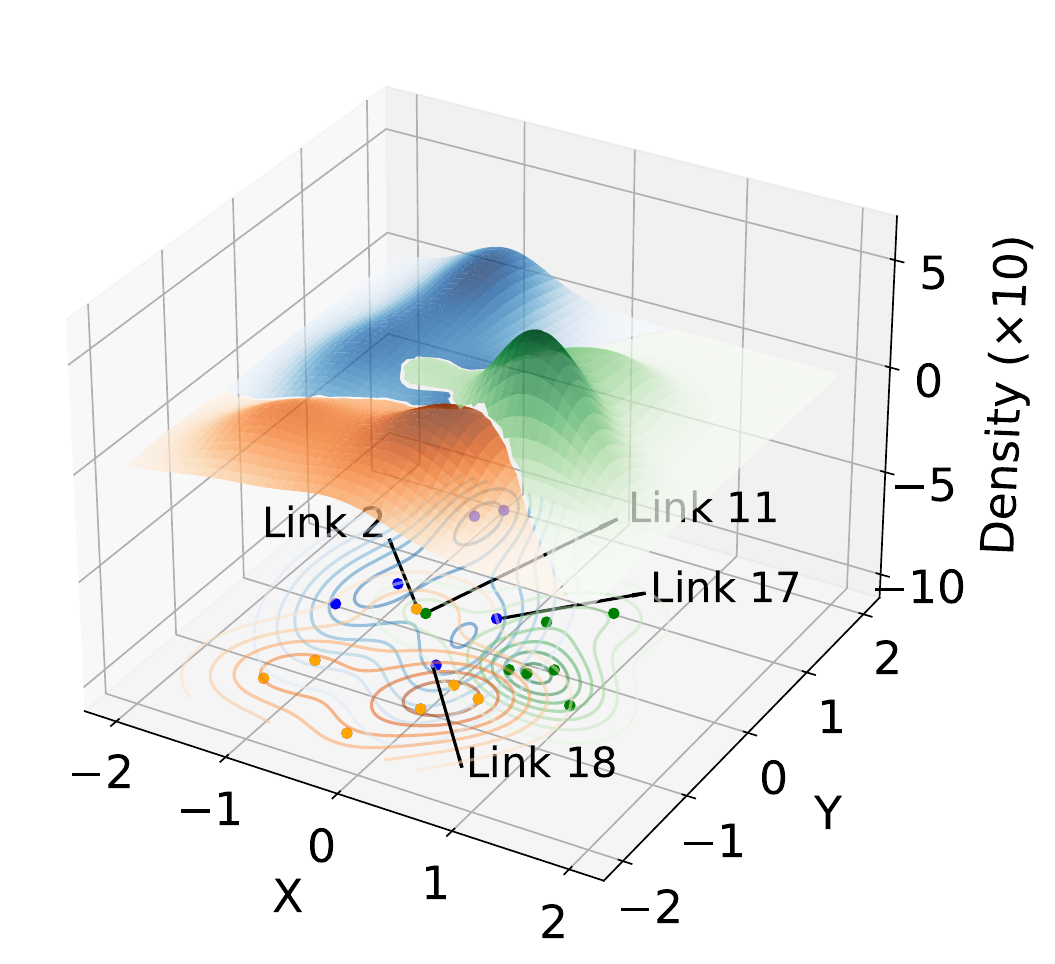}} 
	\subfigure[]{\includegraphics[width=0.24\textwidth]{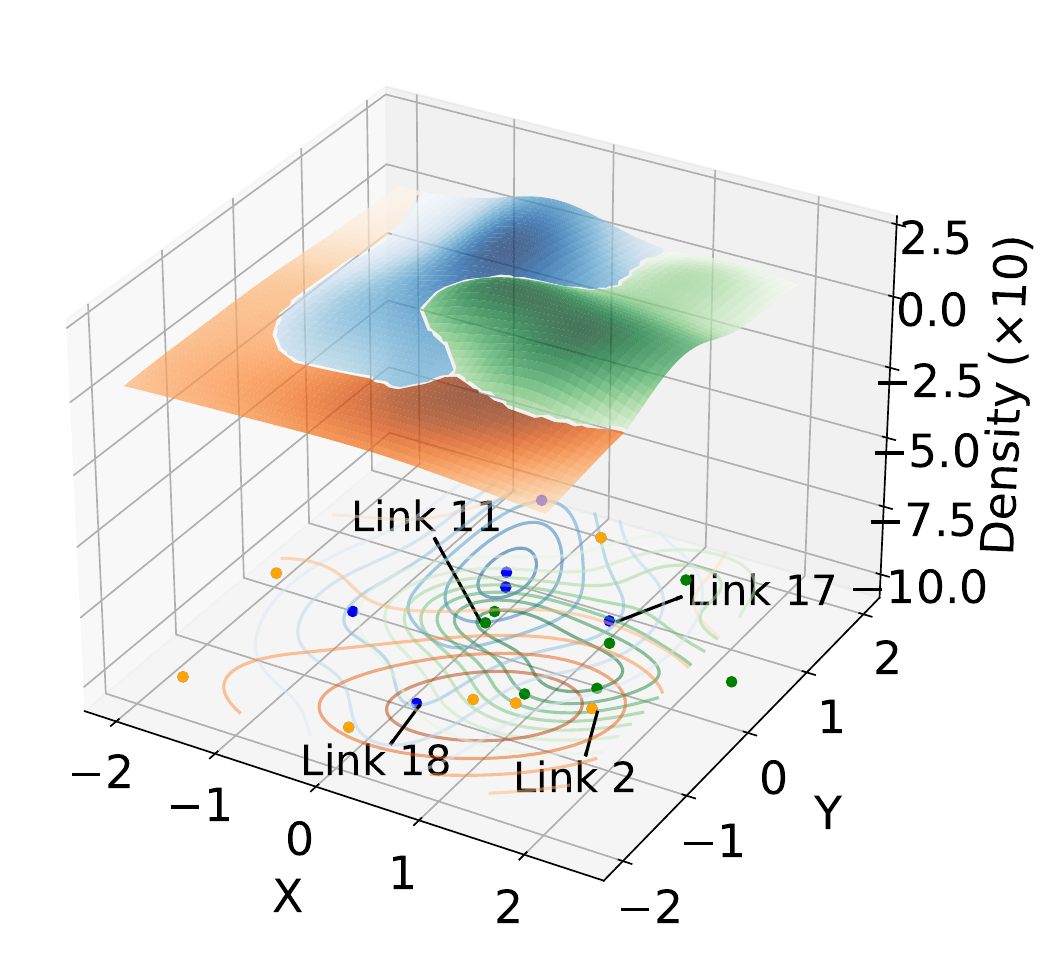}} 
	\subfigure[]{\includegraphics[width=0.24\textwidth]{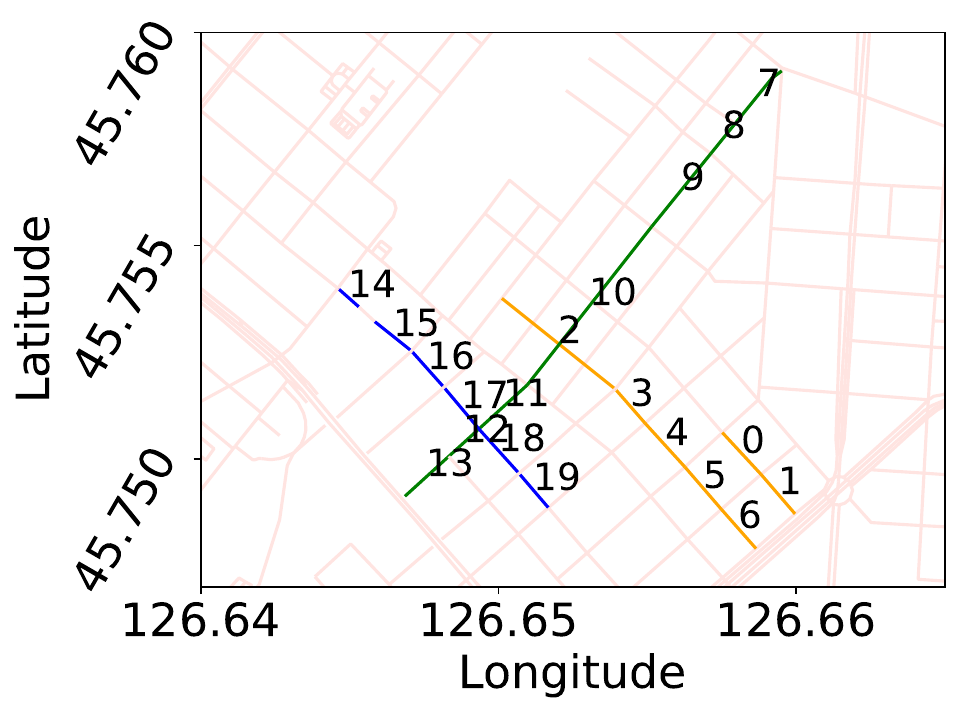}} 
	\caption{Visualization of the projected link representation vectors. (a). 9:00–9:20 AM. (b). 5:00–5:20 PM. (c). 3:00–3:20 AM. (d). The actual locations of the links on the map. The point distributions are derived using kernel density estimation, with density values scaled by a factor of 10 for enhanced clarity.}\label{PCA}
\end{figure}

From the perspective of vector visualization, we selected 20 links with interaction relationships and visualized their representation vectors $\bm{H}^L$ during the morning peak, evening peak, and off-peak periods in two dimensions using Principal Component Analysis (PCA). The distribution of the nodes is depicted using kernel density estimation methods. As shown in Figure~\ref{PCA}, during the morning and evening peak hours, links on the same road exhibit a high degree of clustering, indicating strong correlations. The two-dimensional coordinates of the representation vectors for links near road intersections (such as links 2, 11, 17, and 18) also fall at the boundaries of these clustered categories. In contrast, during off-peak hours, there is a greater overlap between clusters from different roads, and the distribution of points for each road is relatively scattered. This suggests that the correlations among the links become weaker during these periods.

\begin{figure}
	\setlength{\abovecaptionskip}{-0cm}
	\centering
	\subfigure[]{\includegraphics[width=0.24\textwidth]{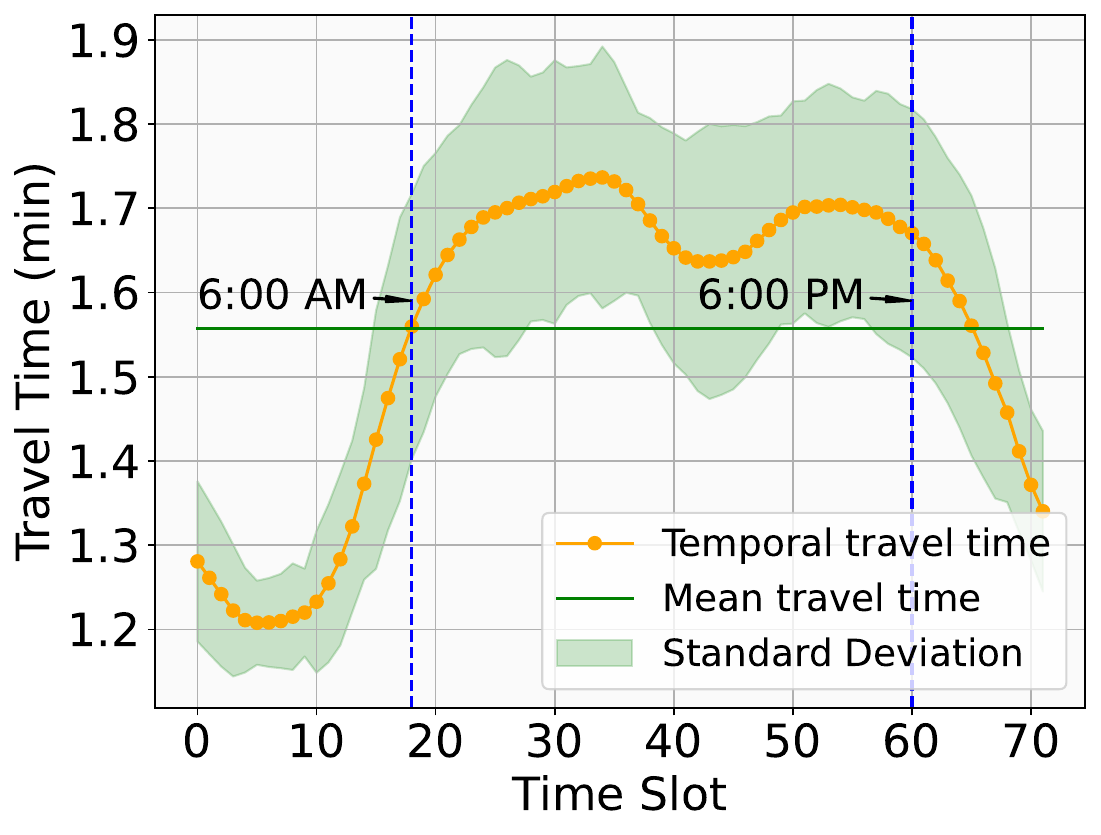}} 
	\subfigure[]{\includegraphics[width=0.24\textwidth]{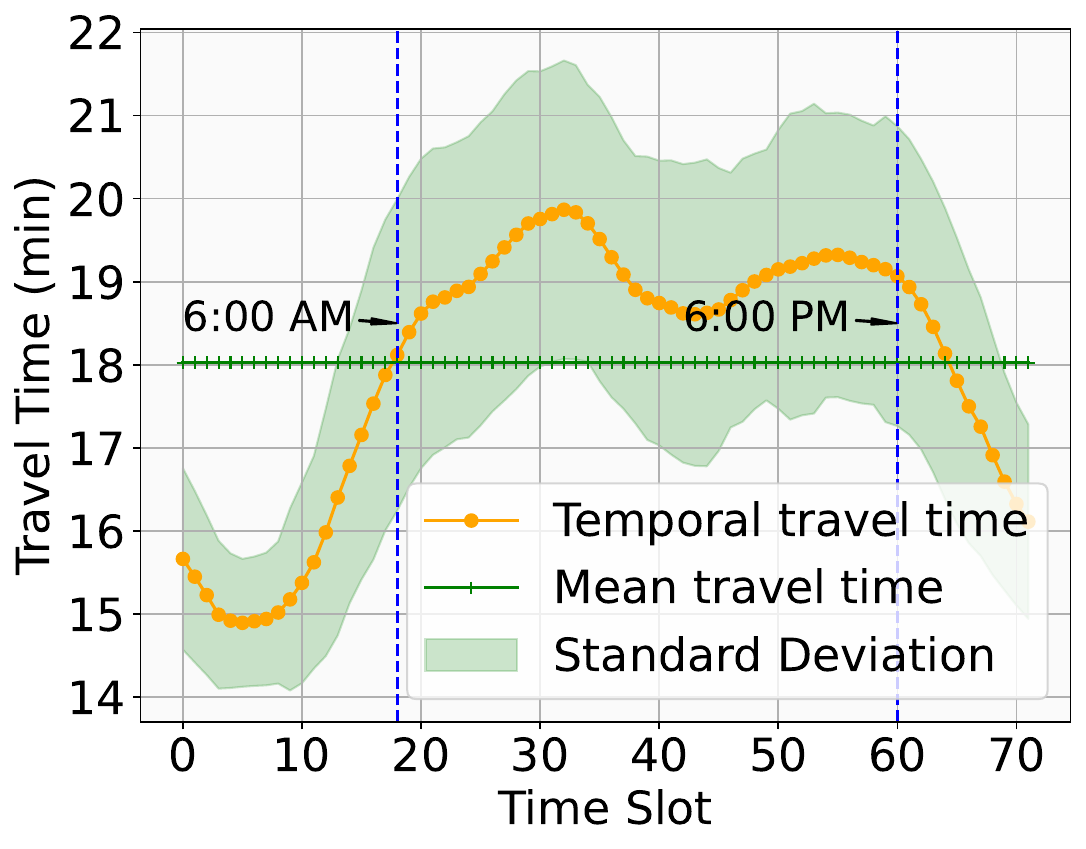}} 
	\subfigure[]{\includegraphics[width=0.242\textwidth]{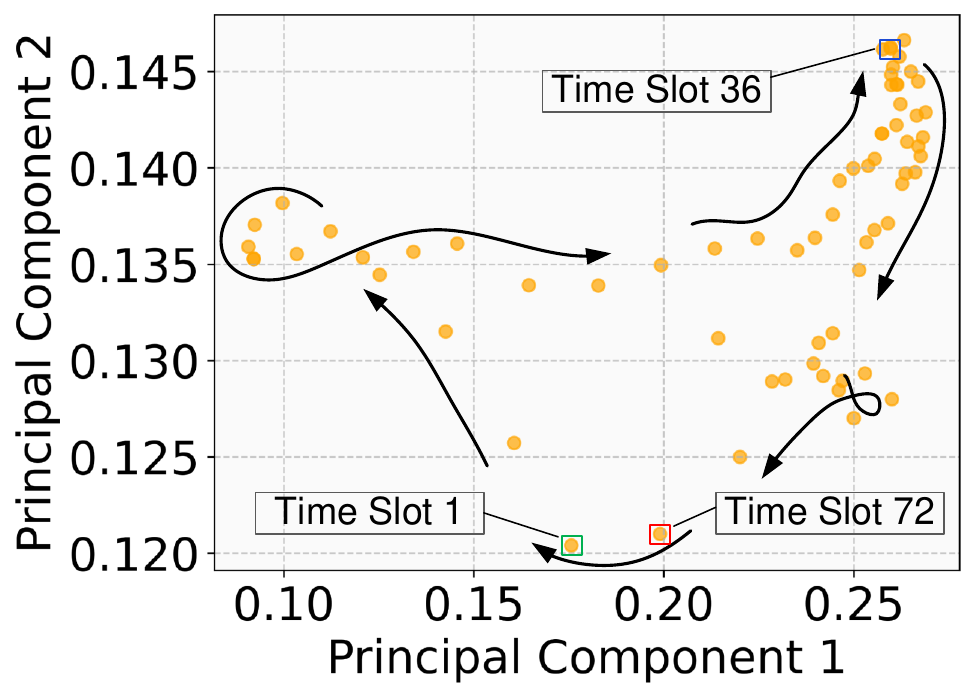}} 
	\subfigure[]{\includegraphics[width=0.238\textwidth]{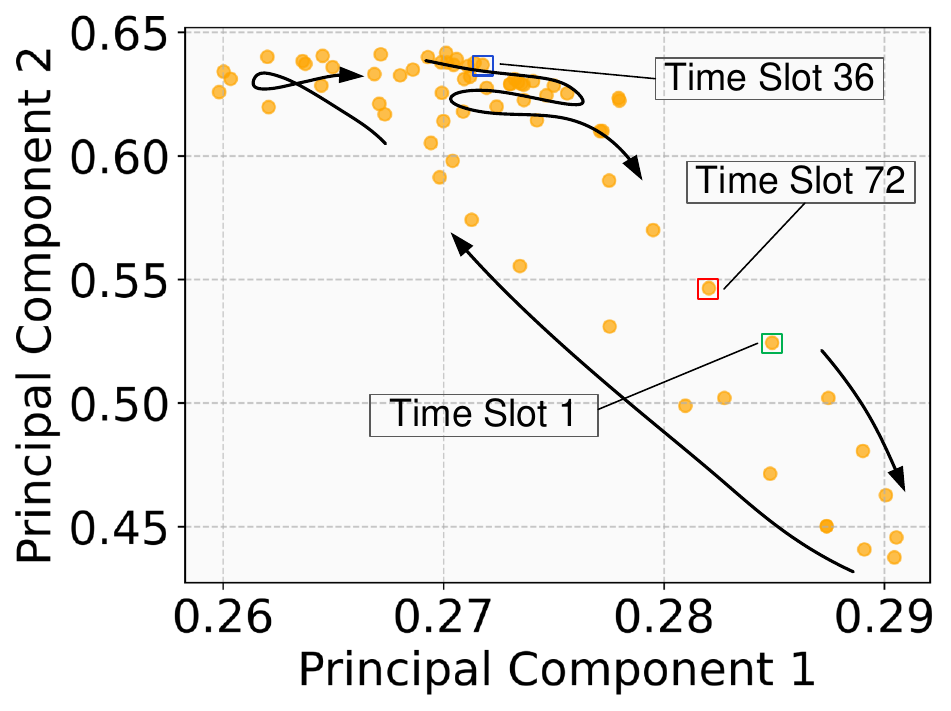}} 
	\caption{Visualization of learned temporal evolution. (a). Temporal evolution of link travel time. (b). Temporal evolution of trip travel time. (c). Temporal evolution of the projection of link mean representation vectors. (d). Temporal evolution of the projection of link correlation representation vectors. }\label{temporalShow}
\end{figure}

Furthermore, we visualized the effectiveness of \textit{SPTTE} in modeling the temporal evolution of travel times. 
We randomly selected 1,000 trips and links and utilized \textit{SPTTE} to output their mean and variance of travel times during different time periods, as shown in Figure~\ref{temporalShow}(a,b). We can find that \textit{SPTTE} effectively models the temporal evolution of travel times for trips and links throughout the day. During off-peak hours, the variance is relatively small, while during peak hours, the variance is larger. Although we cannot obtain the actual travel times for the same trip during different time periods, our time-varying modeling approach demonstrates a superior ability to fit the variations in travel times compared to estimating with a single mean value (green line) throughout the day, resulting in improved predictive performance.

From the perspective of vector visualization, we visualized the time-varying characteristics of the link embedding vectors $\bm{H}^\mu$ (for the mean) and $\bm{H}^L$ (for the correlation) in two dimensions using PCA as shown in Figure~\ref{temporalShow}(c,d). Both the mean and correlation representation vectors exhibit a cyclical trend throughout the day, with the projected positions of adjacent periods showing a high correlation. Based on the learned evolution trends in vector variations, the link representations can be easily refined through interpolation, thereby obtaining mean and covariance values for links and trips at a finer temporal granularity.

\subsection{Analysis of Temporal Evolution Interpolation}

\begin{table*}
\centering
\caption{Interpolation experimental results under different time discretizations.} \small
\begin{tabular}{c c c c c c c c c c c}
\hline

\multirow{2}{*}{\# slots}&\multirow{2}{*}{Interpolation}&  \multicolumn{4}{c}{Chengdu}& &\multicolumn{4}{c}{Harbin}  \\
		 & &  RMSE & MAE &MAPE(\%)& CPRS & &RMSE & MAE &MAPE(\%)& CPRS\\
		\hline
\multirow{2}{*}{72} & --& 134.79& 94.81& 12.25& 1.179&& 157.60& 113.62& 12.96& 1.262\\ 
                             & \checkmark& 135.19& 95.02& 12.25& 1.180&& 157.92 & 113.31& 12.95& 1.262\\ 
\hline
\multirow{2}{*}{24} & --& 135.03& 95.27& 12.31& 1.185&& 158.30 & 114.14& 13.02& 1.267\\ 
                             & \checkmark& 134.96& 95.21& 12.30& 1.184&& 158.16& 113.36& 13.00& 1.264\\ 
\hline
\multirow{2}{*}{12} & --& 135.36& 95.66& 12.36& 1.190&& 159.67 & 114.79& 13.09& 1.275\\ 
                             & \checkmark& 135.09& 95.44 & 12.33& 1.187&& 160.29& 114.55& 13.06& 1.272\\ 
\hline
\multirow{2}{*}{8} & --& 136.02& 96.36& 12.45& 1.198&& 160.70 & 115.59& 13.18& 1.283\\ 
                             & \checkmark& 135.76& 96.07 & 12.41& 1.195&& 159.81& 114.98& 13.11& 1.278\\ 
\hline
\end{tabular}
\label{tb:interploation}
\end{table*}

In the continuous evolution of the link representation over time, the link representation at each time point should carry significance. In \textit{SPTTE}, the link representation vectors are discretely output at uniform time intervals; however, this may not align with the actual occurrence times of the trips. Here we employ linear interpolation to further refine the temporal granularity of the link representations, while also validating the effectiveness of the learned temporal evolution. Specifically, we perform two-step \textit{SPTTE} to output the link representation vector $\bm{H}^s_t,s\in\{\mu,V,L,D\}$ for the current time slot and $\bm{H}^s_{t+1}$ for the next time slot. Then we linearly interpolate between $\bm{H}^s_t$ and $\bm{H}^s_{t+1}$ to obtain the aligned representation vector corresponding to the queried trip's occurrence time and estimate the joint distribution of the trip. We set discrete time intervals as \{20min, 1h, 2h, 3h\}, which correspond to dividing the day into \{72, 24, 12, 8\} time slots, with other experimental conditions held constant. The result is shown in Table~\ref{tb:interploation}.

We observed that temporal interpolation of representation almost always improves performance across different discretization scales, particularly when the discretization scale is large. However, excessively large discretization scales may cause the evolutionary relationship between two representation vectors to no longer fit a linear assumption, which could result in less noticeable improvements or even negatively impact the original performance. Nevertheless, the results provide sufficient evidence that the learned evolution of the link representation vectors is interpretable. This continuous evolution allows the model to construct corresponding representation vectors for trips at any time point, leading to more accurate joint distribution estimates.

\subsection{Analysis of Loss Function and Hyperparameter} \label{sec:vd}

\begin{figure}
	\setlength{\abovecaptionskip}{-0cm}
	\centering
	\subfigure[]{\includegraphics[width=0.238\textwidth]{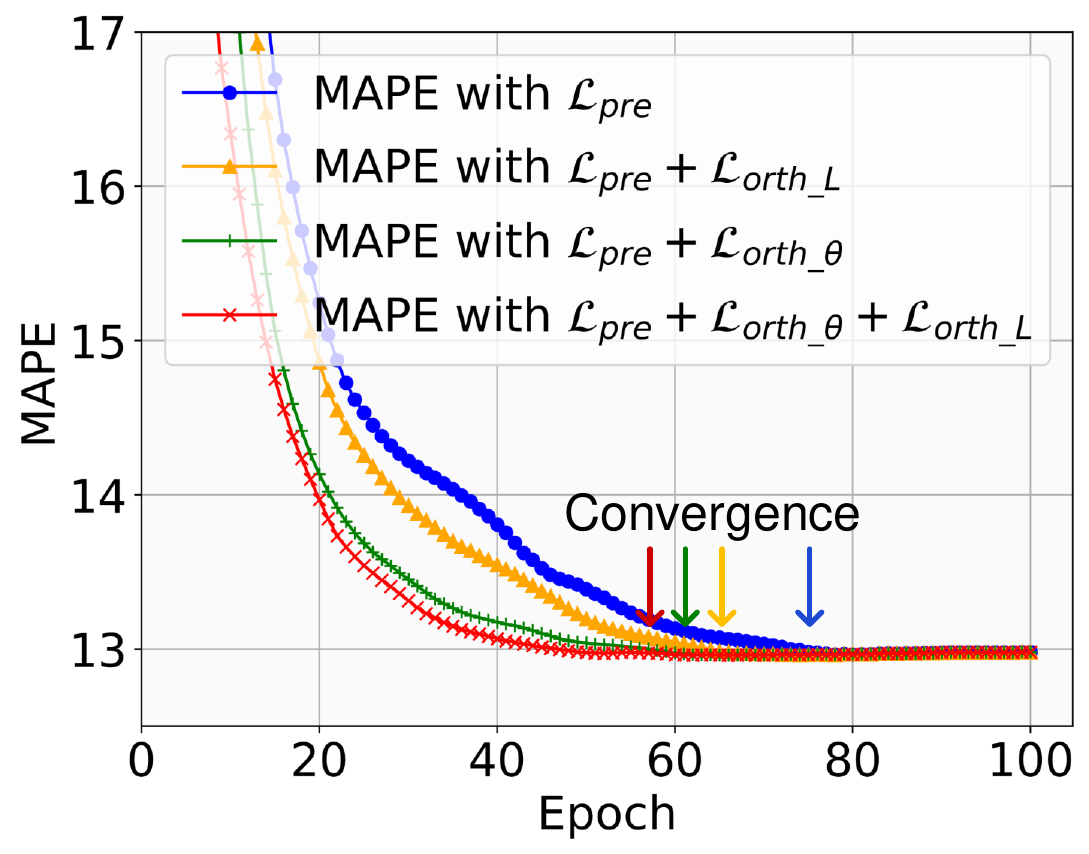}} 
	\subfigure[]{\includegraphics[width=0.242\textwidth]{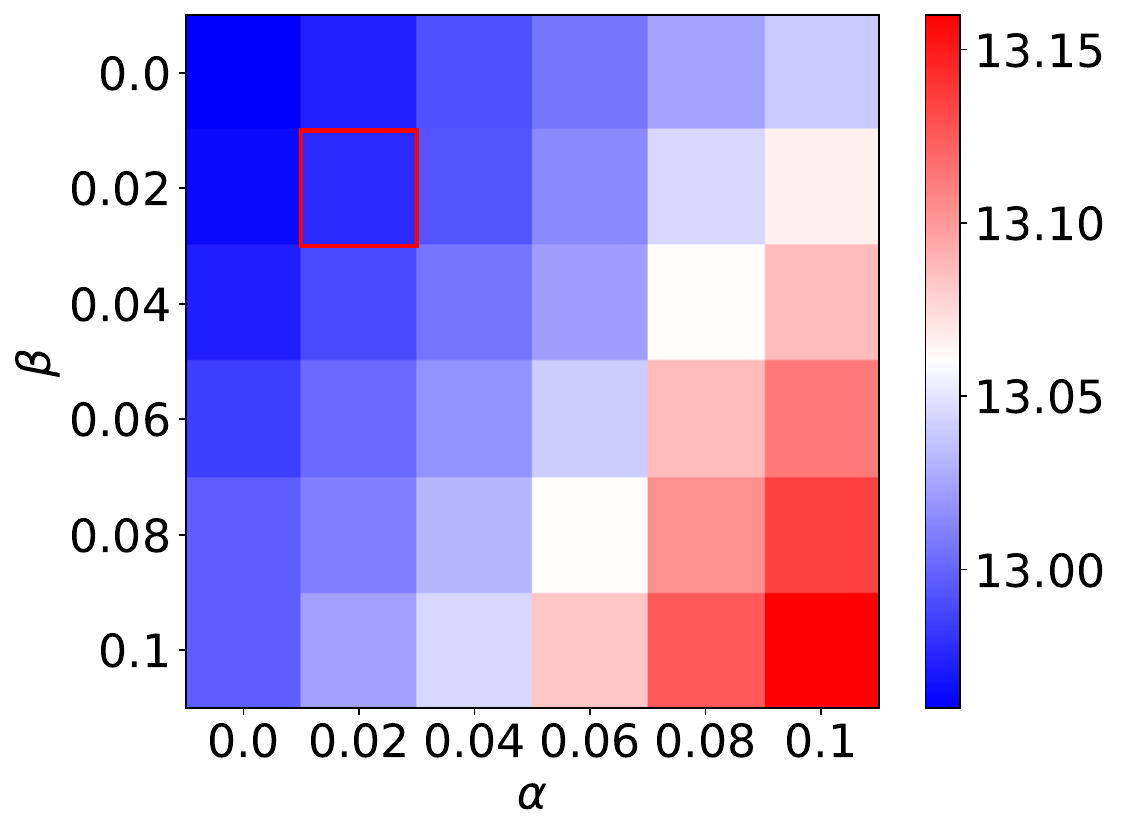}} 
	\caption{Analysis of loss function and hyperparameter. (a). Convergence of \textit{SPTTE} under different loss functions. (b). Grid search for determining hyperparameters.}
	\label{fig:loss}
\end{figure}

We tested the effects of two orthogonal losses in \textit{SPTTE} on the Harbin dataset. We recorded MAPE during the training process under different combinations of losses to illustrate their impact on convergence speed. As shown in Figure~\ref{fig:loss}(a), the orthogonal constraints on both the parameters $\bm{\theta}$ of $f_h(\cdot)$ and the representation vector $\bm{L}$ facilitate the model's convergence speed. Among these, the orthogonal constraint on the parameters exhibits a more pronounced effect. In this case, the model converged at the 76th epoch under the unconstrained setting. The $\mathcal{L}_{\text{orth}_L}$ constraint enables the model to achieve optimal performance approximately 10 epochs earlier, with a 13.16\% improvement, while the $\mathcal{L}_{\text{orth}\_\theta}$ constraint accelerates convergence by about 14 epochs, with a 18.42\% improvement. Together, both constraints result in a combined acceleration of approximately 18 epochs, with a 23.68\% improvement.

$\mathcal{L}_{\text{orth}_L}$ and $\mathcal{L}_{\text{orth}\_\theta}$ serve as two types of regularization constraints. Although they can accelerate model training, the weights associated with them must be meticulously designed; otherwise, they may impose adverse limitations on model performance. We employed a grid search method to evaluate the impact of different hyperparameters $\alpha$ and $\beta $ on model performance. We defined the ranges for $\alpha$ and $\beta$ as $[0, 0.1]$ with a grid spacing of 0.01 and tested the performance of the model on the Harbin dataset with various combinations of $(\alpha, \beta)$. As shown in Figure~\ref{fig:loss}(b), when $\alpha,\beta = 0.02$, the model incorporates the largest possible regularization constraints while maintaining optimal performance. As the hyperparameters increase, the model's performance is gradually negatively affected, with $\alpha$ being more sensitive than $\beta$ and having a greater impact on performance.

\subsection{Analysis of Time Complexity}

\begin{table*}\small
	\centering
	\small
	\caption{Runtime (s) of models.}
	\label{tb:runtime}
	\begin{tabular}{c c c c c c c c } 
		\hline
		{Model}& DeepTTE& HierETA &MulT-TTE&DeepGTT& GMDNet& {ProbETA}& \textbf {SPTTE} \\
		\hline
		{Chengdu }  & { $70.24$} & {$104.91$ }&$153.30$&$44.16$&{$115.66$ }&{${92.31}$ }& $\textbf{193.51}$\\
		
		{Harbin } & { $199.67$} &{$349.50$ }& {$491.45$ }& {{$158.91$} }& $425.37$&{${315.64}$ }&$\textbf{566.18}$\\
		\hline
	\end{tabular}
\end{table*}

We analyze the time complexity based on the four-layer structure of \textit{SPTTE} and its three loss functions. The time complexity of the temporal layer is primarily caused by the GRU. Considering a $k_{gru}$-layer GRU with input and output dimensions of $1$, a hidden dimension of $h$, and recurrence over $\eta$ time steps, the time complexity of this layer is $O(\eta k_{gru}h^2)$. The hidden branching layer divides the hidden state (dimension $h$) and the static embedding (dimension $h$) into four parts through four matrix transformations, resulting in a time complexity of $O(h^2)$. The time complexity of the spatial smoothing layer arises from four $k_{gcn}$-layer GCNs with both input and output dimensions are $2h$ is $O(k_{gcn}|\mathcal{V}|^2h)$. The primary time complexity of the aggregating output layer arises from matrix transformations, which include the link aggregation operation and construction of covariance matrix with a time complexity of $O(b(k_{aug}+1)|\mathcal{V}|h) +b^2(k_{aug}+1)^2h)$, where $b*(k_{aug}+1)$ is the batch size. Among the three loss functions, the highest time complexity is associated with the computation of the maximum likelihood estimation for the travel times of $b*(k_{aug}+1)$ trips. As mentioned above, based on the Woodbury matrix identity and the matrix determinant lemma, its time complexity is $O(b(k_{aug}+1)^2h)$. Considering that $\eta, k_{gru},k_{gcn},k_{aug}, b \ll |\mathcal{V}| $, the overall time complexity is $ O(\eta k_{gru}h^2)+O(h^2)+ O(k_{gcn}|\mathcal{V}|^2h) +O(b(k_{aug}+1)|\mathcal{V}|h) +b^2(k_{aug}+1)^2h) +O(b(k_{aug}+1)^2h)=O(|\mathcal{V}|^2h)$. 
We also compared the actual runtime of our model with that of the baselines. We recorded the time required to train and perform inference for one epoch for each model on the Chengdu and Harbin datasets, utilizing an NVIDIA Tesla V100 GPU for execution. The runtime results are presented in Table~\ref{tb:runtime}. The results demonstrate that our model achieves exceptional performance while maintaining acceptable computational overhead. The increased computational cost is primarily attributed to the calculation of the joint distribution likelihood.

\section{Conclusion} \label{sec:conclusion}

In this paper, we proposed \textit{SPTTE}, a spatiotemporal joint probability estimation model for multi-trip travel time. We formulated the joint distribution estimation of multiple trip travel times as a fragmented observation-based spatiotemporal stochastic process regression problem. Our model captures the time-varying nature of link travel time distributions by leveraging trip coverage frequency characteristics and parameterizing the joint distribution in a low-rank manner based on Gaussian processes. Utilizing the affine transformation of Gaussian distributions, we characterize trips and construct a joint probability distribution for multiple trips. We optimize link representations by maximizing the likelihood of trip travel times, serving as the loss function. To address instability in link representations caused by uneven trip coverage, we heterogeneously introduce prior similarity among links. Experimental results demonstrate that our approach effectively learns interpretable link representations and significantly improves travel time estimation performance, surpassing state-of-the-art baselines by more than 10.13\%. Visualization analyses show that these representations exhibit reasonable temporal variability and spatial similarity. Furthermore, the model's computational complexity and actual training time remain comparable to the baselines.

\bibliographystyle{IEEEtran}
\bibliography{reference}
\end{document}